\documentclass{article} % For LaTeX2e
\usepackage[final]{flageval}
\usepackage{times}
\usepackage{hyperref}
\usepackage{url}
\definecolor{cvprblue}{rgb}{0.21,0.49,0.74}
\definecolor{darkblue}{rgb}{0, 0, 0.5}
\hypersetup{colorlinks=true, citecolor=darkblue, linkcolor=darkblue, urlcolor=darkblue}
\usepackage{amsmath,amsfonts,bm}

\def\eqref#1{equation~\ref{#1}}
\def\1{\bm{1}}

\DeclareMathAlphabet{\mathsfit}{\encodingdefault}{\sfdefault}{m}{sl}
\SetMathAlphabet{\mathsfit}{bold}{\encodingdefault}{\sfdefault}{bx}{n}

\usepackage{longtable}
\usepackage{hyperref}
\usepackage{url}
\usepackage{graphicx}
\usepackage{booktabs}
\usepackage{makecell}
\usepackage{booktabs}
\usepackage{array}
\usepackage{adjustbox}
\usepackage{tabularx}
\usepackage{tcolorbox}
\usepackage{enumitem}
\usepackage{extarrows}
\usepackage{wrapfig}
\usepackage{multirow}

\title{Beyond Multiple Choice: Verifiable OpenQA for Robust Vision-Language RFT}
\author{Yesheng Liu$^{1, 2, 3}$, Hao Li$^{3, 4}$, Haiyu Xu$^{3,5}$, Baoqi Pei$^{6}$, Jiahao Wang$^{1, 2, 3}$, Mingxuan Zhao$^{3,5}$, \\ \textbf{Jing-Shu Zheng}$^{3}$ , \textbf{Zheqi He}$^{3}$, \textbf{JG Yao}$^{3}$, \textbf{Bowen Qin}$^{3}$, \textbf{Xi Yang}$^{3}$, \textbf{Jiajun Zhang}$^{1,2}$ \\
    $^{1}$Institute of Automation, CAS, $^{2}$School of Artificial Intelligence, UCAS,
    $^{3}$BAAI \textit{FlagEval} Team, \\ 
    $^4$BUAA, $^{5}$PKU, $^{6}$ZJU \\
    Project Page: \url{https://flageval-baai.github.io/ReVeL/}
}

\fancyheadoffset[R,L]{0cm}

\begin{document}

\maketitle
\begin{abstract}
Multiple-choice question answering (MCQA) has been a popular format for evaluating and reinforcement fine-tuning (RFT) of modern multimodal language models. Its constrained output format allows for simplified, deterministic automatic verification.
However, we find that the options may leak exploitable signals, which makes the accuracy metrics unreliable for indicating real capabilities and encourages explicit or implicit answer guessing behaviors during RFT. 
We propose ReVeL (\textbf{Re}write and \textbf{Ve}rify by \textbf{L}LM), a framework that rewrites multiple-choice questions into open-form questions while keeping answers verifiable whenever possible.
The framework categorizes questions according to different answer types, apply different rewriting and verification schemes, respectively. When applied for RFT, we converted 20k MCQA examples and use GRPO to finetune Qwen2.5-VL models. Models trained on ReVeL-OpenQA match MCQA accuracy on multiple-choice benchmarks and improve OpenQA accuracy by about six percentage points, indicating better data efficiency and more robust reward signals than MCQA-based training.
When used for evaluation, ReVeL also reveals up to 20 percentage points of score inflation in MCQA benchmarks (relative to OpenQA), improves judging accuracy, and reduces both cost and latency. We will release code and data publicly.

\end{abstract}    
\section{Introduction}
As large language and multimodal models \citep{Anthropic2024Claude4, GPT-5, Bai2025Qwen25VLTR, openai2023gpt4v, Google2025Gemini25pro, Chen2025InternVL2.5, Pei2025EgoThinker} increasingly tackle diverse real-world tasks, the demand for reliable and scalable evaluation has grown significantly. MCQA is convenient because restricting outputs simplifies scoring \citep{Moore2023AssessingMCQ} across language \citep{Hendrycks2020MMLU, Wang2024MMLUPro} and vision language benchmarks \citep{Yue2023MMMU, Yue2024MMMUPro, Liu2023MMBench, Kembhavi2016AI2D, Zhang2024MMERealWorld, Hao2025EMMA}.

However, MCQA departs from real‑world usage where answers are usually open-ended \citep{Lyu2024BeyondPro}, while the predefined options encourage selection heuristics \citep{Zheng2023LMSelectBias, Balepur2024ArtifactsOA} rather than genuine understanding.
%Our analyses show that this format is fragile. We designed 
To quantify the unreliability of MCQA for evaluation and verification, we conduct multiple experiments:
% to quantify this fragility: when options are added to originally open datasets, accuracy rises markedly; when options are replaced with `None of the above' or perturbed, model behavior degrades.
(1) When options are added to the questions in an open-form benchmark, the accuracy metrics can be greatly boosted; (2) In MCQA benchmarks, When the ground-truth option is perturbed, or replaced with `None of the above', model behavior degrades.
These patterns indicate that the MCQA metrics are heavily dependent on the option set, rather than solely on the knowledge and skills required in the question stem.
%This fragility matters because a large share of visual reasoning datasets used MCQA for outcome based RFT.
This fragility matters because many visual reasoning datasets used for outcome-based RFT have included large proportions of MCQA data.
We find that training on MCQA increases multiple-choice accuracy metrics but hurts open-form generalization, widening the gap between the two evaluation settings. In other words, this reward encourages shortcuts tied to options rather than transferable knowledge or reasoning (See Figure \ref{fig:mcqa-fragility}).

Therefore, we present ReVeL (Rewrite and Verify by LLM), a unified framework that rewrites MCQA into open‑ended QA (OpenQA) and preserves verifiability whenever possible. 
%ReVeL triages items into numeric, keyword, per‑option verification, and genuinely generative cases. The first three are judged by deterministic rules; only the last relies on LLM-Judge.
ReVeL categorizes the original multiple-choice questions into numeric, keyword, per‑option verification, and genuinely generative cases. The first three types can be accurately graded by deterministic rules, and only the last type may need an LLM Judge for grading.
This hybrid design reduces cost and variance from a trivial solution that entirely uses an LLM judge for all problems, while maintaining reliability during evaluation.
Across four benchmarks, 70–96\% of items become rule‑verifiable, reaching higher judging accuracy numbers than entirely using a strong LLM judge (GPT 4.1 mini).

Based on ReVeL, we also rewrite 20k MCQA examples into OpenQA and perform GRPO‑based RFT on Qwen2.5‑VL‑3B/7B. Models trained with ReVeL‑OpenQA match MCQA accuracy on choice benchmarks while improving OpenQA accuracy by about six percentage points, demonstrating higher data efficiency and stronger robustness than MCQA‑based training. With a modest data, OpenQA‑trained 7B models also exceed the counterparts trained on open‑source data recipes such as VL‑Rethinker‑7B \citep{Wang2025VLRethinker}, R1‑OneVision‑7B \citep{Yang2025R1Onevision}, and Mixed‑R1‑7B \citep{Xu2025MixedR1} on open‑ended evaluation. In summary, our contributions are threefold:

\begin{figure*}[ht]%[htbp]
\centering
\includegraphics[width=0.9\linewidth]{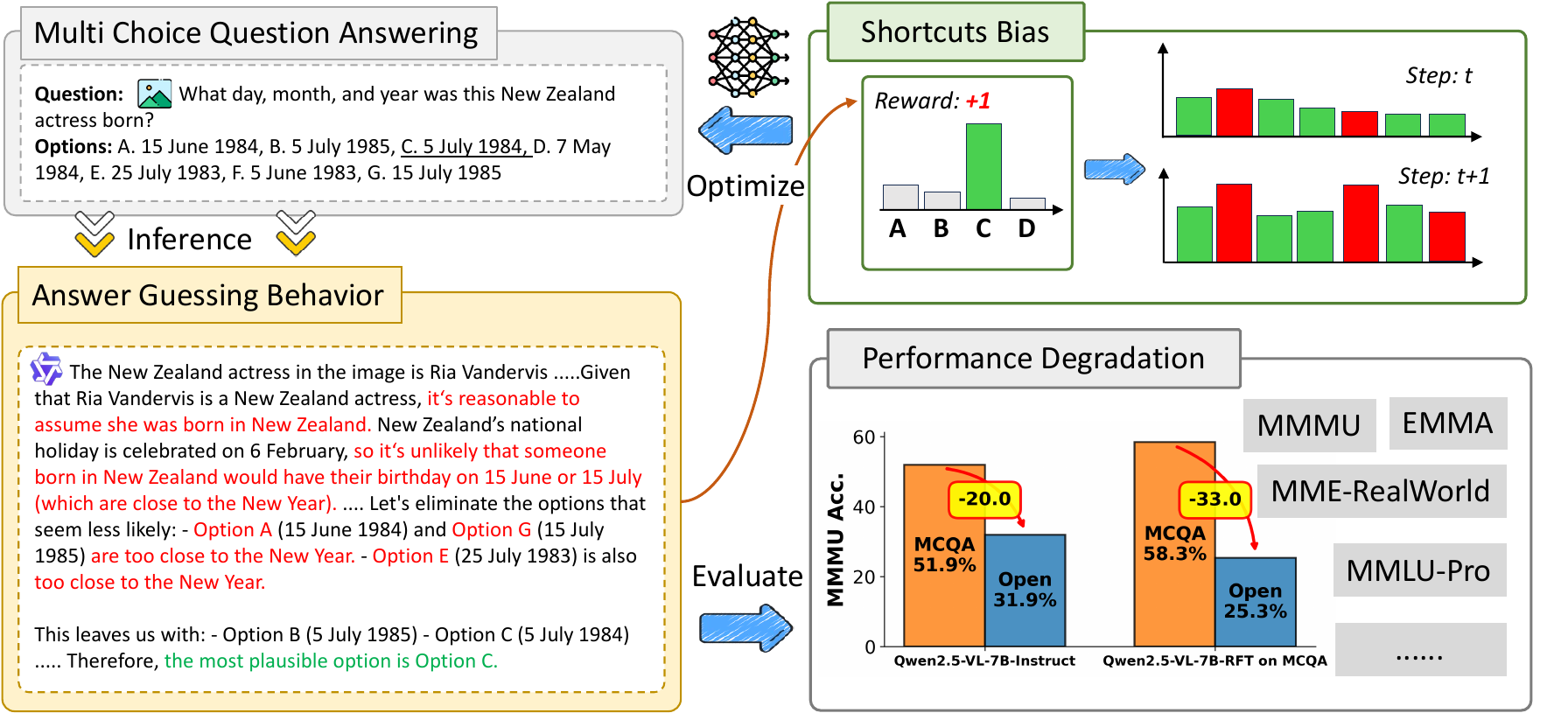}
\caption{Illustration of MCQA fragility. The example (left) shows an unfaithful reasoning chain that eliminates distractors incorrectly yet provide a correct final answer, yielding a positive reward signal that, when used in reinforcement learning, further amplifies shortcut behavior (top right). This shortcut behavior leads to widening gap between MCQA and OpenQA. The diagram motivate us to propose ReVeL, which aligns evaluation and training with reliable OpenQA.}
\label{fig:mcqa-fragility}
\end{figure*}

\begin{itemize}
    \item \textbf{Quantifying the non-robustness of MCQA}: We find that evaluation via MCQA not only makes benchmark scores overestimating true capabilities, but also lacks robustness to trivial modifications of the options. Furthermore, RFT on MCQA improves multiple-choice accuracy at the cost of harming open-ended generalization.
    \item \textbf{The ReVeL framework}: We propose a scalable framework to rewrite MCQA into OpenQA, using accurate rule-based judging whenever possible, with much less cost and variance than entirely shifting to an LLM judge.
    \item \textbf{Demonstration of impact on training and evaluation}: Performing RFT on 20K rewritten samples (Qwen2.5-VL-3B/7B) maintains MCQA accuracy while improving OpenQA accuracy by ~6 percentage points. Rewriting four benchmarks also reveals up to 20 percentage points of score inflation when shifting from MCQA to OpenQA.
    % \item We will release code, rewritten datasets, and evaluation scripts to support reliable, reproducible research.
\end{itemize}

\section{Fragility of MCQA}
\label{sec:mcqa-analysis}

% We show that the multiple choice format systematically suffers from fragility. We organize the evidence into: (i) accuracy increases when options are added to originally open datasets rises accuracy; (ii) performance degrades when an answer is replaced by `None of the above'; and (iii) performance drops after options are removed. We then connect these findings to how MCQA bias reinforcement finetuning.
Our work is directly motivated by a series of experiments that quantitatively expose the weaknesses of the MCQA format.
We describe our methodology and results here.
% Our work is directly motivated by a series of experiments that expose the weaknesses of the MCQA format via quantitative analysis.
% We describe our methodology and results in this section.

\subsection{Adding options to open-ended benchmarks}
%\subsection{OpenQA is much simpler if options added}
\begin{figure}[ht]
    \centering
    \centering
    \includegraphics[width=0.65\linewidth]{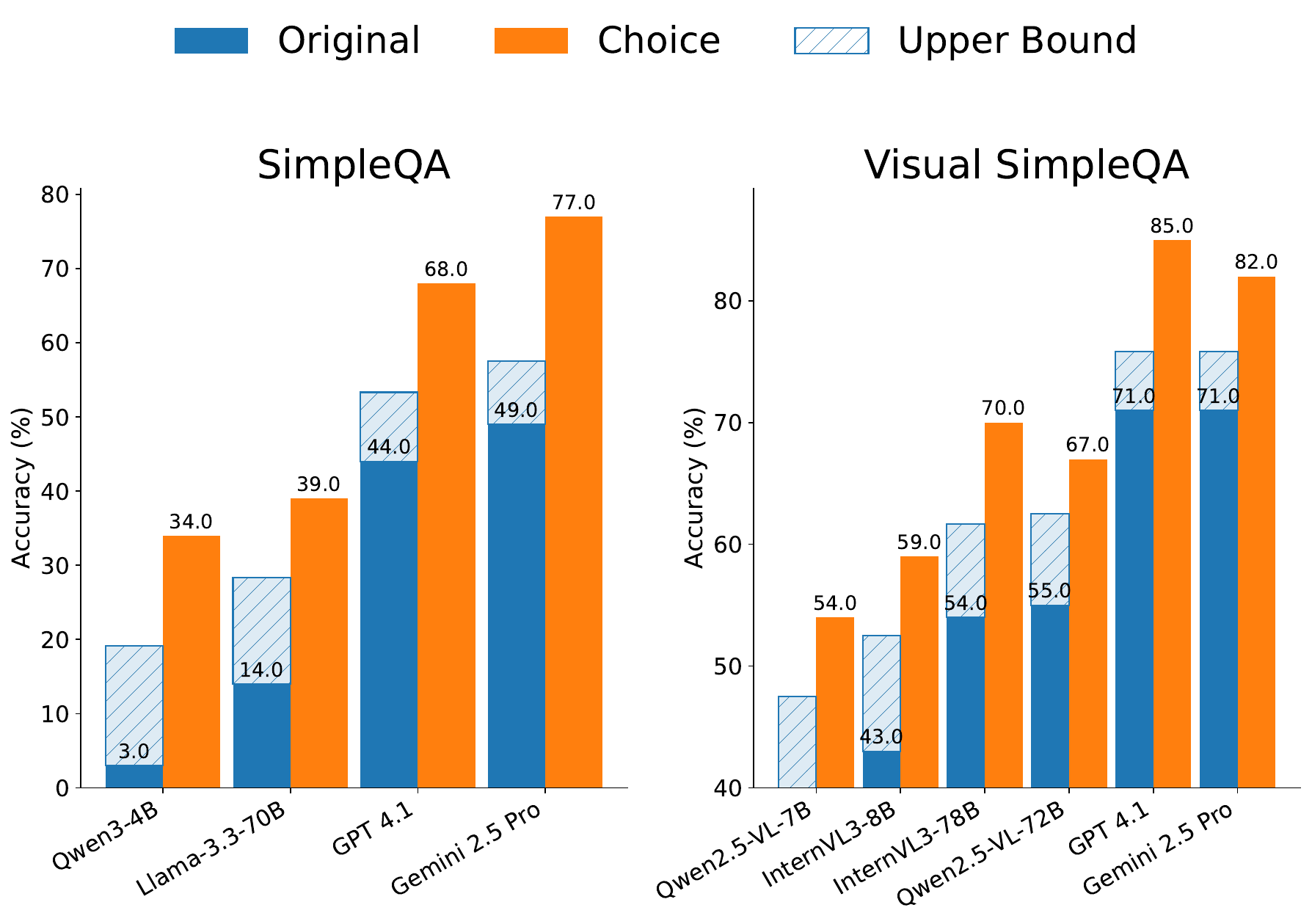}
    \caption{Performance comparison on original open-ended datasets (SimpleQA, Visual SimpleQA) and their multiple-choice versions (*-Choice, with 6 options). The \textit{Random Guess} score is a theoretical upper bound that combines the model's actual open-ended accuracy with the probability of correctly guessing on the rest of the questions from six options.}
    \label{fig:open2mcq}
\end{figure}

\textbf{Setup.} We start from two 
%established open‑ended QA datasets:
recent benchmarks that expect free-form answers from an LLM or VLM:
SimpleQA~\citep{Simpleqa} and VisualSimpleQA~\citep{Wang2025VisualSimpleQA}.
%, where the native tasks require free‑form answers without any pre-specified set of options.
%To isolate the effect of options, we 
We convert each question into an MCQA variant (SimpleQA‑Choice / VisualSimpleQA‑Choice) by retaining the ground-truth answer and adding five plausible distractors via a human‑in‑the‑loop procedure with GPT‑4.1.
%,using prompts from \citet{Zhang2025AutomatedOpen2Choice}.
This conversion preserves the original semantics, but the metrics may be affected by random guessing.
% while making elimination and chance guessing possible. Accuracy is the primary metric. To contextualize ceilings due purely to guessing, we report a random‑guessing upper bound accuracy metric as
Therefore, besides accuracy, we also report a random‑guessing upper bound:
\[
Acc_{UB} = Acc_{Open} + (1 - Acc_{Open}) \times \frac{1}{K}, K=6
\]
i.e., the model answers correctly on items it can already solve in open‑ended form and guesses uniformly on the rest.

\textbf{Findings.} Across both open‑weight (e.g., Qwen2.5‑72B, Llama‑3.3‑70B) and proprietary models (e.g., GPT‑4.1, Gemini 2.5 Pro), converting to MCQA yields consistently large gains relative to the open‑ended baseline and the random‑guessing upper bound (Figure \ref{fig:open2mcq}). This pattern holds for both text‑only (SimpleQA) and multimodal (VisualSimpleQA) settings, indicating that 
%models benefit from information embedded in the options rather than improved underlying reasoning.
when a model correctly answers a multiple-choice question, it is often utilizing the information embedded in the option set even when it does not actually have the required knowledge or reasoning skills.

\textbf{Implication.} The presence of options supplies huge extra signal that can be exploited independent of task competence, directly leading to
%benchmark inflation under MCQA.
overestimation of model capabilities from MCQA accuracy.

\subsection{Replacing GT with None-of-the-Above}
Another way to test the target knowledge or reasoning skill is to replace the ground-truth option with an option to abstain: `None of the above (is correct)' (NOTA), after shifting the remaining false options frontwards.
%We next employ `None of the Above' (NOTA) on the MMLU-Pro and MMMU datasets to further diagnose the fragility of multiple-choice evaluation
We conduct such an experiment on MMLU-Pro~\citep{Wang2024MMLUPro} and MMMU~\citep{Yue2023MMMU}, the most popular MCQA benchmarks for LLM and VLM evaluation.
%(See benchmark selection strategy in \ref{sec:benchmark}).
\begin{figure}[htbp]
\centering
\includegraphics[width=0.8\linewidth]{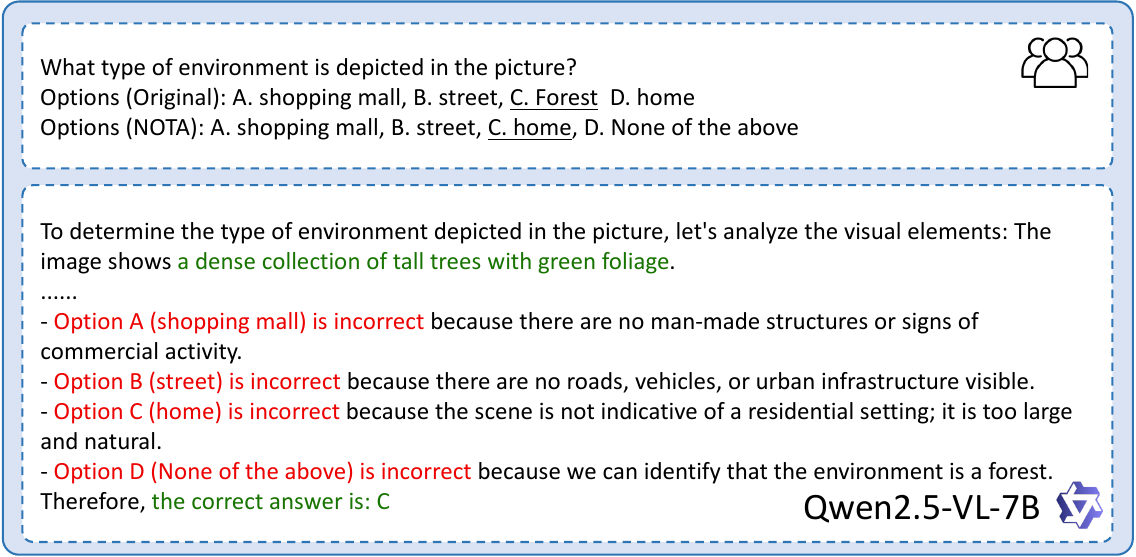}
%\caption{Illustration of the Reasoning-answer Mismatch. The model's reasoning deny the option C, but choose it as the final choice}
\caption{Reasoning and answer can mismatch after replacing the ground-truth option with NOTA.}
\label{fig:mismatch}
\end{figure}

When the correct option is replaced by NOTA, models frequently display 
%internal inconsistencies: the chain-of-thought correctly rejects the true answer yet still selects an eliminated option.
a logical inconsistency: the chain-of-thought reasoning process sometimes correctly eliminates the incorrect options yet still selects one of them as the final answer.
As shown in Figure \ref{fig:mismatch}, such contradictions occur even when the model explicitly reasons towards the correct concept (``forest'' in that example) but finalizes with an inconsistent choice (``C. home'').
Quantitatively, mismatch rates rise from 18\% in standard MCQA to 50\% under NOTA, listed in Appendix
% \ref{tab:mismatch_results}.
%A second failure mode, positional memorization, reveals that
Also, we notice that models often reuse the original ``correct'' letter position even after the content was modified (after shifting) henceforth incorrect ( listed in Appendix
%\ref{tab:positional_memorization}), 
implying potential test set contamination or shallow recall of positional cues.
%rather than semantic verification. 

Together, these effects expose how fragile MCQA could be, motivating the shift to option-free OpenQA evaluation.

\subsection{Omitting the options from an MCQ}
\label{sec:mcqa2openqa}
To examine the genuine reasoning ability without the aid of options, we can also 
%convert multiple-choice questions into open-ended questions.
remove the options for some multiple-choice questions, treating them as open-form questions.
Note that after removing the options, some questions are still valid,
% through a filtering pipeline. 
but some would become ill-posed.\footnote{For instance, ``How many apples are in the basket?'' is still a valid question without any options, but ``Which of the following statements are true?'' is not. We illustrate four primary categories of questions that cannot apply option removal in the supplementary appendix.}
%Therefore, we discard questions exhibiting such unsuitability patterns or yielding identical answers after option removal. As a result, 
Based on an LLM-assisted analysis (prompt attached in appendix), we find that
only about half of the questions in widely used MCQA benchmarks remain suitable 
% for open-ended conversion:
using open-form evaluation:
48.9\% for MMLU-Pro and 44.1\% for MMMU, shown in Table \ref{tab:open_suitable}.

\begin{table}[htbp]
\centering
\small
\caption{Proportion of open-ended questions after filtering.}
\begin{tabular}{l|c|c}
\toprule
Dataset & Total & Open Ratio (\%) \\
\midrule
MMLU-Pro (sampled) & 1000 & 48.9 \\
MMMU (validation) & 900 & 44.1 \\
% \midrule
% Total & 1900 & 46.6 \\
\bottomrule
\end{tabular}
\label{tab:open_suitable}
\end{table}

\begin{figure}[ht]
    \centering
    \includegraphics[width=0.8\linewidth]{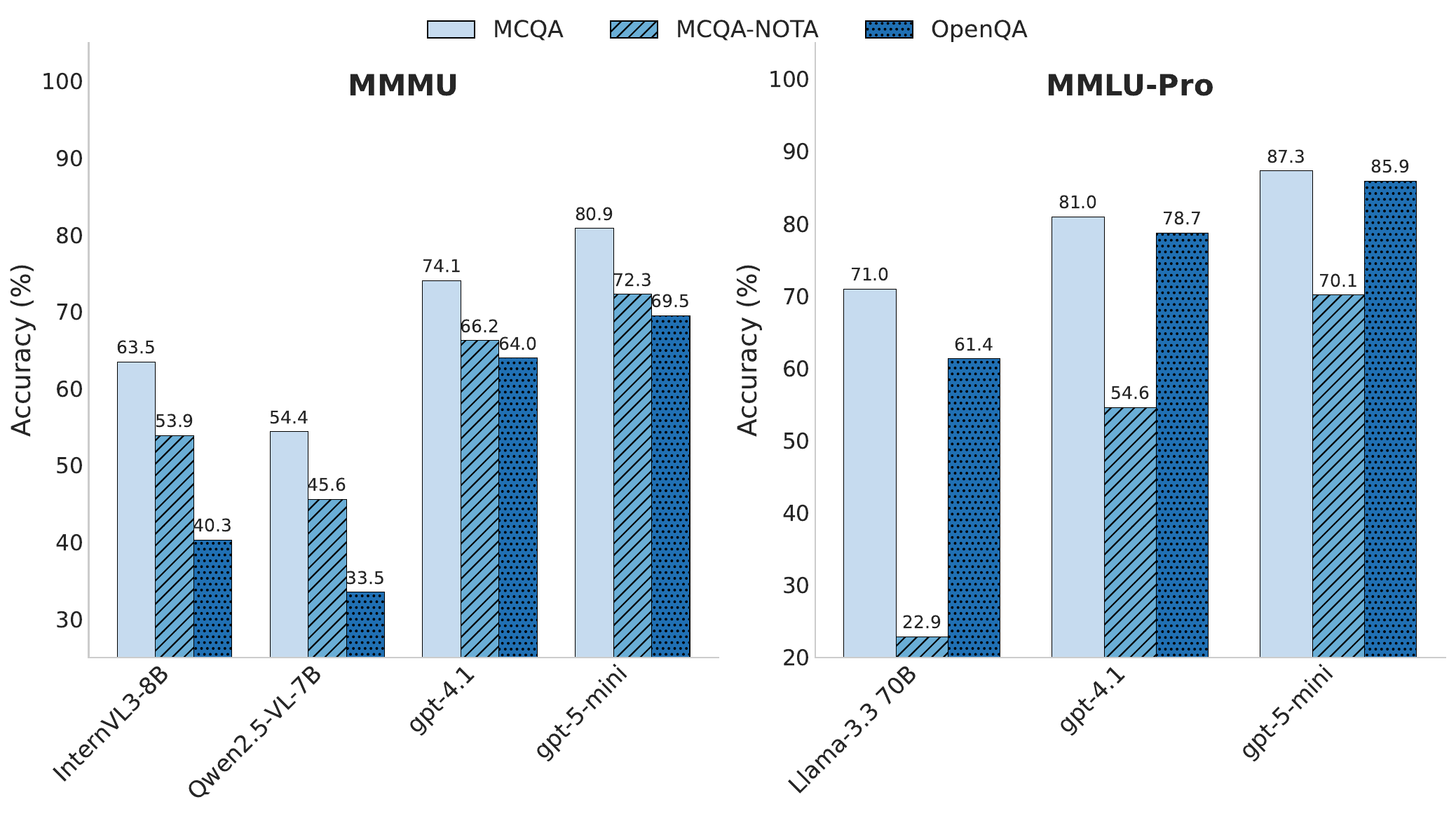}
    % \caption{Performance comparison on original multiple choice datasets (MMMU, MMLU-Pro) and their open ended versions (Open). Across most models, accuracy is significantly higher when options are provided.}
    \caption{On the impact of options on multiple-choice benchmarks: when options are removed, accuracy is uniformly lower, especially on VQA benchmarks like MMMU.}
    \label{fig:mcq2open_dif}
\end{figure}

%On this filtered subset, 
On the same questions that are still valid without options,
%removing options and posing the problems in open form 
%open-form evaluation leads to consistent accuracy degradation across models and datasets.
models achieve consistently lower accuracy than the original MCQA format,
as shown in Figure \ref{fig:mcq2open_dif}.
%, confirming that MCQA inflate evaluation. The following sections quantify this inflation and analyze its implications for MCQA-based training.

\subsection{RFT on MCQA hurts open-ended QA}
\label{sec:mcqa_rft}
Finally, we study training effects by utilizing reinforcement fine-tuning on MCQA data and evaluating on both MCQA and their open counterparts described in Section \ref{sec:mcqa2openqa}. We use the popular GRPO algorithm \citep{Shao2024DeepSeekMath} in this work for RFT experiments. RFT on MCQA improves MCQA scores but degrades open-ended performance, thereby widening the MCQA–OpenQA gap. For example, on MMMU, the gap grows for both 3B and 7B models; similar trends hold on EMMA (see Table \ref{tab:combined_model_comparison}). This indicates that the verifiable reward under MCQA may overfit to option-specific heuristics rather than transferable reasoning.

\begin{table}[h]
\centering
\small
\label{tab:rft-results}
\caption{\textbf{Impact of RFT on ViRL MCQA data.}
MCQ = multiple-choice benchmark score; Open = Open-ended benchmark score.  
$\Delta$ denotes the inflation gap (MCQ–Open).  
RFT on ViRL (5K MCQA samples) improves MCQ scores but enlarges $\Delta$, indicating reinforced shortcut behavior.
}
\begin{tabular}{cccc}
\toprule
\textbf{Model} & \textbf{MCQA} & \textbf{OpenQA} & \textbf{$\Delta$ (Acc Drop)} \\
\midrule
\multicolumn{4}{c}{\textit{MMMU}} \\
\midrule
Qwen2.5-VL-3B & 46.6 & 11.8 & 34.8 \\
\quad + MCQA (ViRL) & 50.9 & 11.6 & 39.3 \textcolor{cvprblue}{( +4.5)} \\
\midrule
Qwen2.5-VL-7B & 51.6 & 21.4 & 30.2 \\
\quad + MCQA (ViRL) & 56.4 & 17.1 & 39.3 \textcolor{cvprblue}{( +9.1)}\\
\midrule
\multicolumn{4}{c}{\textit{MMLU-Pro}} \\
\midrule
Qwen2.5-VL-3B & 39.5 & 21.1 & 18.4 \\
\quad + MCQA (ViRL) & 47.4 & 20.4 & 27.0 \textcolor{cvprblue}{( +8.6)} \\
\midrule
Qwen2.5-VL-7B & 53.4 & 27.6 & 25.8 \\
\quad + MCQA (ViRL) & 53.6 & 27.0 & 26.6 \textcolor{cvprblue}{( +0.8)} \\
\bottomrule
\end{tabular}
\label{tab:combined_model_comparison}
\label{tab:filter_model_comparison}
\end{table}

Across settings, MCQA enables option exploitation that inflates accuracy,
%(ii) lacks semantic invariance under option removal or NOTA,  and (iii) induces reasoning answer mismatches. 
amplifies shortcuts tied to options during training. These findings motivate our Rewrite‑and‑Verify approach in Section \ref{sec:hybrid}, which mitigate these shortcuts for both evaluation and training.
% decouples evaluation from option‑level shortcuts while preserving scalability.

\section{ReVeL: The Rewrite-and-Verify framework}
\label{sec:hybrid}
We have shown that MCQA suffers from several shortcomings both in evaluation and in providing reliable training signals. Transforming MCQA to open-ended QA (OpenQA) has the potential to address these issues.
In this work, we introduce \textbf{ReVeL} (\textbf{Re}write-and-\textbf{Ve}rify by \textbf{L}LMs), a framework that rewrites MCQA into open ended yet verifiable formats while ensuring semantic fidelity and minimizing information loss.

\subsection{Pipeline overview}

\begin{figure}[htbp]
\centering
\includegraphics[width=0.8\linewidth]{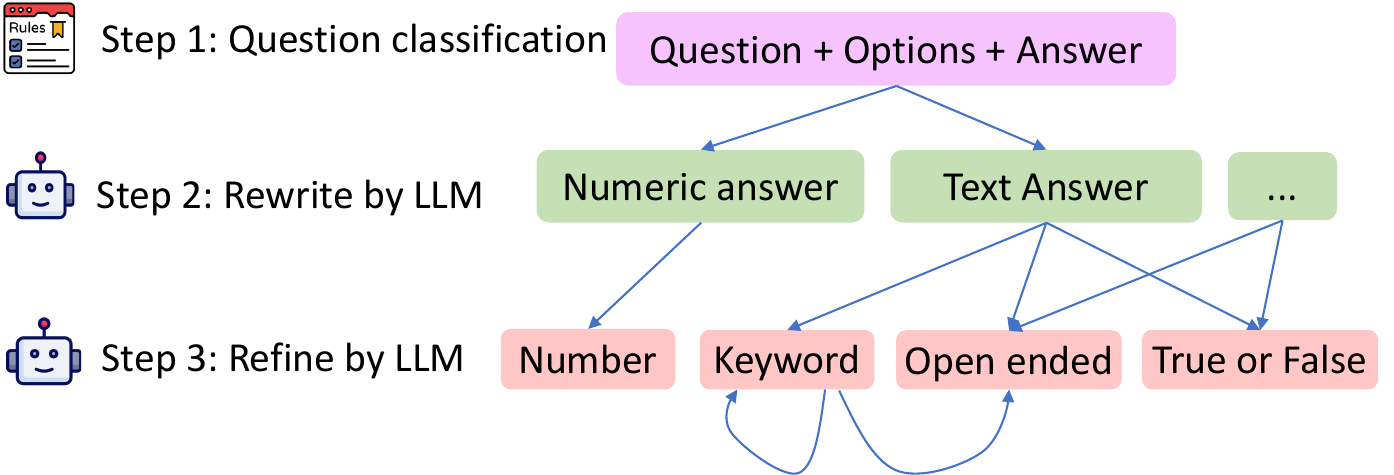}
\caption{Illustration of the rewrite-and-verify framework}
\label{fig:rewrite_pipeline}
\end{figure}

As summarized in Figure \ref{fig:rewrite_pipeline}, ReVeL operates in three phases: (1) Triage and Classification, (2) Prompt-based Rewriting, and (3) Hybrid Evaluation and Verification.
The core principle is to maximize deterministic, rule-based evaluation for questions with unambiguous answers, while reserving LLM-based judging only for cases that genuinely require semantic understanding.

During Triage, questions are first passed through a rule-based filter to leave out those expecting numeric answers,
%(e.g., quantities, ratios, or values with units). 
mostly quantities or ratios such as \textit{50kg} or \textit{$9.8 \times 10^{-23} m/s^2$}.
These will be processed via pattern matching.
Remaining non-numeric questions are routed to a lightweight LLM-assisted classifier that assigns each question to one of three answer verification categories:%semantic categories:
\begin{itemize}
    \item Keywords matching: single or short tokens that have limited variations (e.g., names, dates).% unambiguous
    \item Open answers: short, factual or descriptive sentences that are unambiguous for a typical human or LLM grader.%requiring generation.
    \item Per-option verification: questions heavily depend on the option set, such as \textit{Which of the following statements describes the process of ...}.%discriminative items whose correctness depends on verifying each option (e.g., “Which of the following…”).
\end{itemize}

Each category is paired with a tailored rewriting prompt with the goal to preserve semantics while enabling deterministic verification. Examples of all four categories and their rewritten counterparts are shown in Figure \ref{tab:hybrid_evaluation_examples}
\begin{itemize}
    \item \textbf{Numeric}. ReVeL reformulates them into explicit quantitative prompts by incorporating measurement units and specifying answer format (e.g., comma separated or value–unit pairs).
    \item \textbf{Keywords}. The rewriting step enumerates acceptable synonyms or lexical variants to permit flexible but rule consistent matching.
    \item \textbf{Open answers}. These are rephrased into concise free form queries that solicit factual, non subjective responses without relying on the original options.
    \item \textbf{Per-option verification}. Each option is converted into a declarative statement, and models output a comma separated list of True/False judgments, enabling structured verification and preserving the discriminative intent of MCQA.
\end{itemize}

\subsection{Benchmarks and rewriting coverage}
\label{sec:benchmark}

We evaluate ReVeL on four major multimodal benchmarks, including EMMA, MMMU, MME-RealWorld and MMLU-Pro. \textbf{EMMA}~\citep{Hao2025EMMA} targets multimodal reasoning in STEM, emphasizing visual–textual integration; we focus on the physics and chemistry subsets for domain-specific evaluation. \textbf{MMMU}~\citep{Yue2023MMMU} assesses college-level, multi-discipline reasoning across six domains with diverse image types; we use its 900-question validation set. \textbf{MME-RealWorld}~\citep{Zhang2024MMERealWorld} offers large-scale, high-quality, real-world tasks with greater difficulty; we adopt its “Lite” subset of 1,700 questions. \textbf{MMLU-Pro}~\citep{Wang2024MMLUPro} is a more challenging variant of MMLU, incorporating reasoning-oriented questions, ten-choice answers, and cleaner data. We sample 1,000 questions for evaluation.

\begin{table}[htbp]
\centering
\small
\caption{Performance comparison of hybrid pipeline versus entirely using an LLM judge}
\label{tab:error_comparison}
\begin{tabular}{l@{\hspace{4pt}}lcccc}
\toprule
Dataset & Judger & {Recall $\uparrow$} & {PPV $\uparrow$} & {FPR $\downarrow$} & {Acc. $\uparrow$} \\
\midrule
\multirow{2}{*}{EMMA}   & LLM  & 100 & 100 & 0.0 & 100 \\
                        & ReVeL & 100 & 100 & 0.0 & 100 \\
\midrule
\addlinespace
\multirow{2}{*}{MME-RW} & LLM  & 93.5 & 98.6 & 1.4 & 95.9 \\
                        & ReVeL & 95.7 & 100 & 0.0 & 98.0 \\
\midrule
\addlinespace
\multirow{2}{*}{MMLU-Pro} & LLM  & 95.1 & 97.5 & 3.2 & 95.8 \\
                          & ReVeL & 100 & 100 & 0.0 & 100 \\
\midrule
\addlinespace
\multirow{2}{*}{MMMU}   & LLM  & 100 & 95.0 & 5.4 & 97.3 \\
                        & ReVeL & 93.2 & 98.6 & 1.3 & 96.0 \\
\midrule
\multirow{2}{*}{Overall} & LLM  & 96.4 & 97.2 & 2.0 & 97.3 \\
                         & ReVeL & \textbf{96.8 }& \textbf{99.6} & \textbf{0.3} & \textbf{98.5} \\
\bottomrule
\end{tabular}
\end{table}

\subsection{Judge accuracy and efficiency}
\begin{table}[t]
\centering
\small
\caption{\textbf{Evaluation format distribution after rewriting.} 
``Num'', ``Text'', and ``Opt'' denote rule-based deterministic categories, while ``Open'' requires LLM judging. The large fraction of rule-based items demonstrates the efficiency of our hybrid evaluation design comparing to pure LLM-judge. }
\label{tab:eval_distribution}
\begin{tabular}{lcccc}
\toprule
\multirow{2}{*}{\textbf{Dataset}} & \multicolumn{1}{c}{\textbf{LLM}} & \multicolumn{3}{c}{\textbf{Rule-based}} \\
\cmidrule(lr){2-2} \cmidrule(lr){3-5}
 & \textbf{Open(\%)} & \textbf{Num(\%)} & \textbf{Text(\%)} & \textbf{Opt(\%)} \\
\midrule
EMMA           & 4.1  & 39.0 & 6.6  & 50.3 \\
MMMU           & 17.0 & 31.3 & 33.5 & 18.2 \\
MME-RW  & 28.4 & 3.3  & 55.7 & 12.6 \\
MMLU-Pro       & 20.8 & 39.7 & 19.6 & 19.9 \\
\bottomrule
\end{tabular}
\end{table}
To enhance evaluation consistency and efficiency, ReVeL reclassifies the majority of tasks into deterministically verifiable categories: numeric, keyword, and per-option verification. This design substantially reduces both computational cost and subjective variance by eliminating unnecessary LLM judgment on straightforward verifiable cases.

To validate robustness, we compare ReVeL’s hybrid evaluation against a pure LLM-judge baseline across 600 randomly sampled responses from GPT-4.1-mini, Qwen2.5-VL-7B, and Qwen2.5-VL-72B on four benchmarks. As shown in Table \ref{tab:error_comparison}, ReVeL achieves an overall accuracy of 98.5\%, exceeding the LLM judge’s 97.3\%, while simultaneously reducing false positive rate from 2.0\% to 0.3\%. These trends indicate that integrating rule-based verification improves evaluative stability by enforcing stricter decision boundaries and confirms the robustness of the hybrid verification design.

\begin{table*}[htbp]
    \centering
    \small
    \caption{Examples of our ReVeL Pipeline applied to different question types. Each quadrant displays an original multiple-choice question and its OpenQA counterpart.}
    \label{tab:hybrid_evaluation_examples}
    \renewcommand{\arraystretch}{1.2} 
    
    \begin{tabular}{cl}
        \toprule
        
        \begin{minipage}[c]{0.2\textwidth}
            \centering
            \includegraphics[width=0.8\linewidth]{examples/numeric.png}
            \vspace{1ex} \\
            {\large\textbf{Numeric}} 
        \end{minipage}
        & 
        \begin{minipage}[c]{0.7\textwidth} 
            \raggedright
            \textbf{Original:} An ideal vapor-compression refrigeration cycle that uses refrigerant-134a as its working fluid maintains a condenser at 800 kPa and the evaporator at 212°C. Determine this system's COP and the amount of power required to service a 150 kW cooling load. \\
            \textbf{Options:} A. 4.07, 31.8 kW, B. 4.97, 33.8 kW, \underline{C. 4.87, 30.8 kW}
            
            \vspace{1.5em} 
            \textbf{Rewritten Question:} An ideal vapor-compression refrigeration cycle that uses refrigerant-134a as its working fluid maintains a condenser at 800 kPa and the evaporator at 212°C. Determine this system's coefficient of performance (COP) and the amount of power required to service a 150 kW cooling load, in kilowatts. Provide your answer as two numbers separated by a comma: COP, power (kW). \\
            \textbf{Rewritten Answer:} 4.87, 30.8
        \end{minipage}
        \\
        \midrule
        \addlinespace[0.5em] 
        \begin{minipage}[c]{0.2\textwidth} 
            \centering
            \includegraphics[width=0.8\linewidth]{examples/open-ended_resize.png}
            \vspace{1ex} \\ 
            {\large\textbf{Open answer}} 
        \end{minipage}
        &
        \begin{minipage}[c]{0.7\textwidth}
            \raggedright 
            \textbf{Original:} Goya created this work while \\
            \textbf{Options:} A. in political exile in England B. serving as a soldier on the front lines against France \underline{C. working as the court painter to the king of Spain} D. studying Classical antiquity in Rome.
            
            \vspace{1.5em} 
            \textbf{Rewritten Question:} Goya created this work while holding what professional position? \\
            \textbf{Rewritten Answer:} Working as the court painter to the king of Spain
        \end{minipage}
        \\
        \midrule
        \addlinespace[0.5em] % 在样本之间添加垂直间距
        
        % --- 样本 3: Per-option verification ---
        % 第 1 列: 图像和标题
        \begin{minipage}[c]{0.25\textwidth} % 垂直居中
            \centering
            \includegraphics[width=0.9\linewidth]{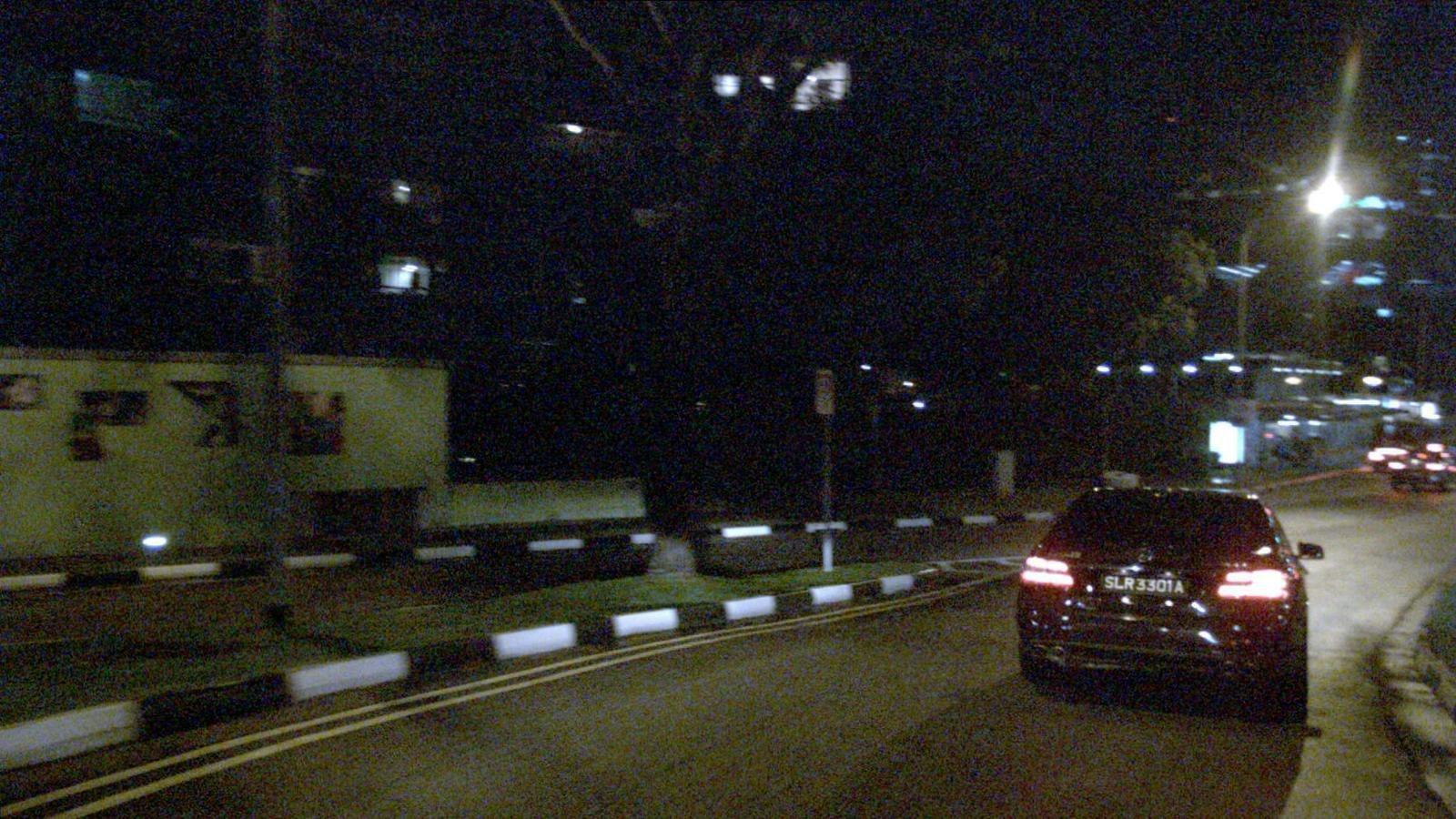}
            \vspace{1ex} \\ % 在图像和标题间添加间距
            {\large\textbf{Per-option verification}} % <-- 标题移到此处
        \end{minipage}
        &
        % 第 2 列: 文本
        \begin{minipage}[c]{0.7\textwidth} % 垂直居中
            \raggedright 
            \textbf{Original:} This image shows the front view of the ego car. Predict the behavior of the ego vehicle. \\
            \textbf{Options:} \underline{(A) The ego vehicle is steering to the right. The ego vehicle is driving fast.} (B)... (C)... (D)... (E)...
            
            \vspace{1.5em} 
            \textbf{Rewritten Question:} This image shows the front view of the ego car. Predict the behavior of the ego vehicle. Now, evaluate each of the following statements about the ego vehicle's behavior. (A)... Provide your answer as a single, comma separated list of True or False values corresponding to statements A through E. \\
            \textbf{Rewritten Answer:} True, False, False, False, False
        \end{minipage}
        \\
        \midrule
        \addlinespace[0.5em] % 在样本之间添加垂直间距
        
        % --- 样本 4: Keywords ---
        % 第 1 列: 图像和标题
        \begin{minipage}[c]{0.25\textwidth} % 垂直居中
            \centering
            \includegraphics[width=0.9\linewidth]{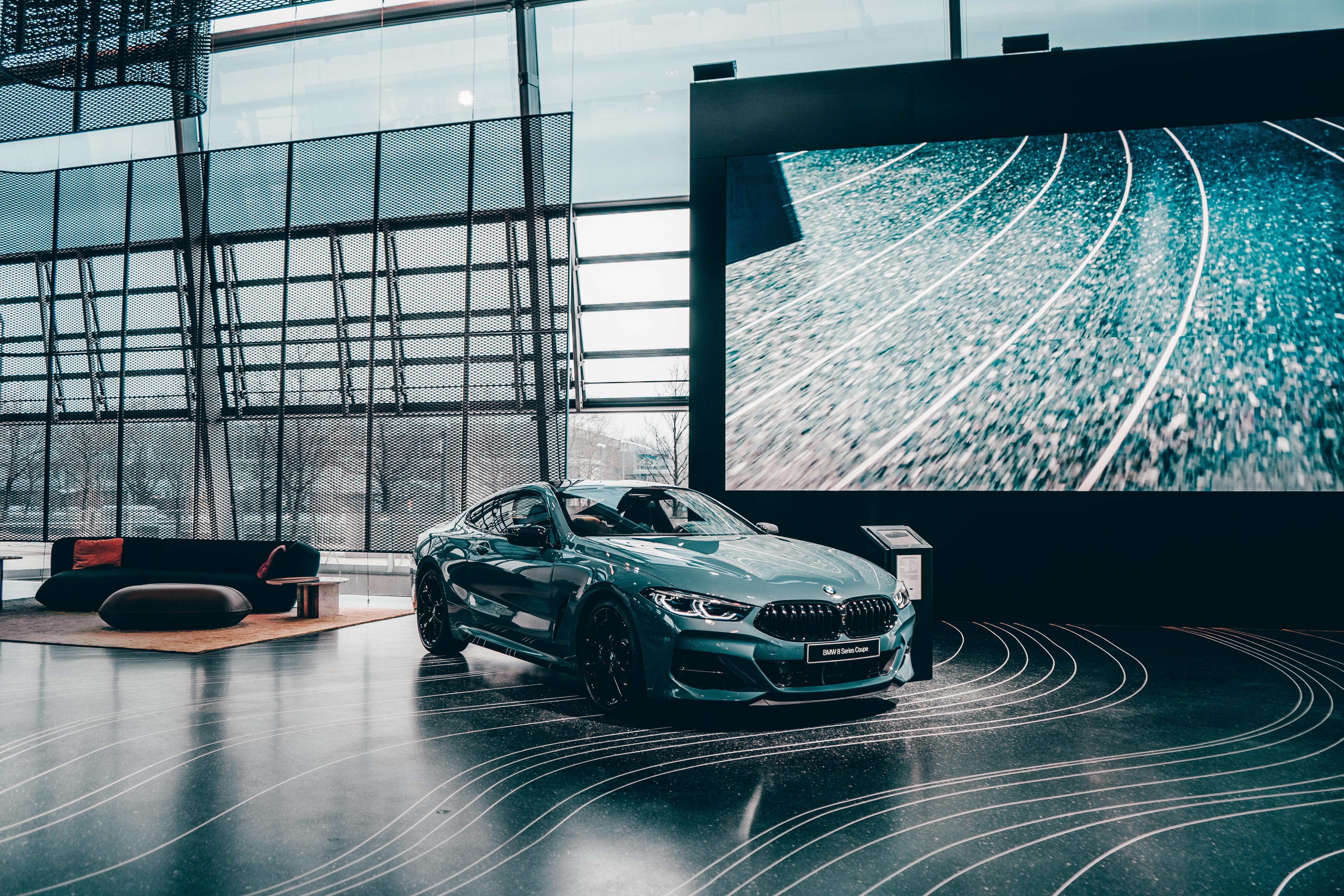}
            \vspace{1ex} \\ % 在图像和标题间添加间距
            {\large\textbf{Keywords}} % <-- 标题移到此处
        \end{minipage}
        &
        % 第 2 列: 文本
        \begin{minipage}[c]{0.7\textwidth} % 垂直居中
            \raggedright 
            \textbf{Original:} What is the manufacturer of the vehicle in the picture? \\
            \textbf{Options:} (A) Mercedes Benz (B) FORD \underline{(C) BMW} (D) HYUNDAI (E) This image doesn't feature the content.
            
            \vspace{1.5em} 
            \textbf{Rewritten Question:} What is the manufacturer of the vehicle in the picture? \\
            \textbf{Rewritten Answer:} BMW $\langle \text{OR} \rangle$ Bayerische Motoren Werke $\langle \text{OR} \rangle$ BMW AG
        \end{minipage}
        \\
        
        \bottomrule
    \end{tabular}
\end{table*}

ReVeL’s rewriting not only improves accuracy but also yields substantial efficiency gains. By turning many open-ended questions into structured formats, most items can now be graded automatically with simple rules. This reduces the need for costly and sometimes inconsistent LLM-based judging. As reported in Table \ref{tab:eval_distribution}, between 70\% and 96\% of questions across datasets can be evaluated through deterministic rules. For example, 95.9\% of EMMA items become fully rule-checkable after rewriting, and even in MME-RealWorld’s complex visual tasks, 71\% are deterministically verifiable. 
\section{Experiments}

\begin{table*}[!h]
  \centering
  \small
  \caption{Performance Comparison of MCQA vs. OpenQA Training on In-Domain and Out-of-domain Benchmarks}
  \begin{tabular}{lrr|rr|rr|rr|ccc}
    \toprule
    \multirow{3}{*}{Model / Train} 
      & \multicolumn{4}{c|}{In-domain} 
      & \multicolumn{4}{c|}{Out-of-domain} 
      & \multicolumn{3}{c}{Overall Scores} \\
    \cmidrule(lr){2-5} \cmidrule(lr){6-9} \cmidrule{10-12}
      & \multicolumn{2}{c}{EMMA} 
      & \multicolumn{2}{c|}{MMMU} 
      & \multicolumn{2}{c}{MME-RW} 
      & \multicolumn{2}{c|}{MMLU-Pro} 
      & \multirow{2}{*}{MCQ} 
      & \multirow{2}{*}{Open} 
      & \multirow{2}{*}{Total} \\
    \cmidrule(lr){2-3} \cmidrule(lr){4-5} \cmidrule(lr){6-7} \cmidrule(lr){8-9}
      & MCQ & Open & MCQ & Open & MCQ & Open & MCQ & Open & & & \\
    \midrule
    R1-Onevision-7B & 28.9 & 4.7 & 42.2 & 23.9 & 44.6 & 31.6 & 42.5 & 32.3 & 39.5 & 23.1 & 31.3 \\
    Mixed-R1-7B & 29.8 & 13.2 & 56.3 & 30.6 & 45.6 & 32.8 & 51.4 & 37.7 & 45.8 & 28.6 & 37.2 \\
    VL-Rethinker-7B & 30.6 & 14.9 & 53.9 & 33.4 & 44.3 & 32.7 & 52.4 & 37.6 & 45.3 & 29.6 & 37.5 \\
        \midrule
    Qwen2.5-VL-3B & 27.4 & 5.7 & 44.3 & 23.3 & 35.9 & 26.6 & 38.7 & 29.6 & 36.6 & 21.3 & 28.9 \\
    \quad + MCQA (ViRL) & 28.2 & 3.1 & 50.2 & 22.0 & 39.7 & 25.6 & 44.0 & 28.0 & 40.5 & 19.7 & 30.1 \\
    \quad \quad + OpenQA (ViRL) & 31.0 & 4.4 & 50.2 & 23.8 & 42.1 & 28.6 & 43.9 & 30.3 & 41.8 & 21.8 & 31.8 \\
    \quad + OpenQA (ReVeL) & 29.8 & \textbf{18.6} & 49.4 & 27.4 & 41.2 & 31.9 & 42.2 & 34.1 & 40.7 & 28.0 & 34.3 \\
    % \quad + MCQA + OpenQA (ViRL) & 31.0 & 4.4 & 50.2 & 23.8 & 42.1 & 28.6 & 43.9 & 30.3 & 41.8 & 21.8 & 31.8 \\
    \quad \quad + OpenQA (ViRL) & \underline{31.4} & \underline{17.3} & 49.4 & 26.5 & 41.4 & 31.7 & 41.3 & 33.4 & 40.9 & 27.2 & 34.1 \\
        \midrule
    Qwen2.5-VL-7B & 28.9 & 10.2 & 51.9 & 31.9 & 44.8 & 32.8 & 49.1 & 39.0 & 43.7 & 28.5 & 36.1 \\
    \quad + MCQA (ViRL) & 30.2 & 9.1 & 58.3 & 25.3 & 50.1 & 32.0 & 52.8 & 32.4 & 47.8 & 24.7 & 36.3 \\
    \quad \quad + OpenQA (ViRL) & \textbf{31.7} & 10.4 & \underline{58.2} & 33.4 & 47.6 & 36.3 & \textbf{53.7} & 37.7 & \underline{47.8} & 29.5 & 38.6 \\
    \quad + OpenQA (ReVeL) & 29.2 & 17.1 & 56.4 & \textbf{37.0} & \textbf{50.6} & \textbf{38.8} & 51.1 & \textbf{43.0} & 46.8 & \textbf{34.0} & \textbf{40.4} \\
    \quad \quad + OpenQA (ViRL) & 29.8 & 16.9 & 54.3 & \underline{36.8} & \underline{50.3} & \underline{38.4} & 51.5 & \underline{39.9} & 46.5 & \underline{33.0} & \underline{39.8} \\
    \bottomrule
  \end{tabular}
  \label{tab:merged-results}
\end{table*}
In this section, we apply our ReVeL framework to rewrite existing visual reasoning datasets for reinforcement learning. Firstly, we find that training with our new data improves both accuracy in MCQA and open-end QA format. Then we use our data for evaluation and observe that there is a large performance gap between MCQA and OpenQA across existing MLLMs.

\subsection{Expertmental settings}

% training details
% We implement all experiments on the VeRL framework with a near on‑policy RL setup and train for up to 10 epochs. We do not use KL regularization. For ViRL‑Open/MCQA‑5K, we use a training batch size of 256, PPO mini‑batch size 128, and rollout size 8. For Mixed‑R1‑Open/MCQA‑15K, we use a training batch size of 512, PPO mini‑batch size 256, and rollout size 8. Inference and serving for all models are done with vLLM. These settings are fixed across regimes to isolate the effect of the reward design.

As discussed in Section \ref{sec:mcqa_rft},  training with MCQA tends to reinforce option-exploiting behaviors and amplify format shortcuts, which can degrade model performance. Thus, we employ ReVeL to convert MCQA datasets into OpenQA form for training. 

% Specifically, we extract the MCQA subsets from ViRL \citep{Wang2025VLRethinker} and Mixed-R1 \citep{Xu2025MixedR1} and get ViRL-MCQA-5K and Mixed-R1-MCQA-15K respectively. Each subset is then rewritten through our ReVeL pipeline into open-ended, verifiable counterparts, yielding ViRL-Open-5K and MixedR1-Open-15K.

% We train Qwen2.5-VL-3B and Qwen2.5-VL-7B with 4 training configurations based on the ViRL dataset: 
We train Qwen2.5-VL-3B and Qwen2.5-VL-7B with GRPO. To conduct a controlled comparison of the impact of different training data, we designed 4 training configurations based on the ViRL dataset, as shown in Table \ref{tab:merged-results}
\begin{enumerate}
    \item \textbf{Original MCQA Only }(\textit{+MCQA (ViRL))}:  The baseline model is trained exclusively on the original ViRL MCQA data. Rewards are derived from rule-based exact match.
    \item \textbf{Original MCQA \& Original OpenQA} (\textit{+OpenQA (ViRL))}: This configuration auguments (1) by further adding the original OpenQA questions from the ViRL dataset. 
    \item \textbf{Rewritten OpenQA Only} (\textit{+OpenQA (ReVeL))}: The baseline model is trained exclusively on the OpenQA data rewritten by our ReVeL pipeline.
    \item \textbf{Rewritten OpenQA \& Original OpenQA} (\textit{+OpenQA (ViRL))}: This configuration augments (3) by further adding the original OpenQA questions from the ViRL dataset.
\end{enumerate}
This setup enables a controlled comparison between reinforcement driven by MCQA versus OpenQA by our ReVeL. Our evaluation is based on the four benchmarks mentioned above.

\subsection{Training details}
We implement all experiments on the VeRL framework with a near on‑policy RL setup and train for up to 10 epochs. We do not use KL regularization. For ViRL‑Open/MCQA‑5K, we use a training batch size of 256, PPO mini‑batch size 128, and rollout size 8. For Mixed‑R1‑Open/MCQA‑15K, we use a training batch size of 512, PPO mini‑batch size 256, and rollout size 8. Inference and serving for all models are done with vLLM. These settings are fixed across regimes to isolate the effect of the reward design.

\subsection{Performance on rewritten training data}

As shown in Table \ref{tab:merged-results}, training on OpenQA consistently produces hight overall accuracy than MCQA across both model sizes: Qwen2.5-VL-3B achieves 34.3 overall with OpenQA vs 30.1 with MCQA (+4.2), and Qwen2.5-VL-7B achieves 40.4 vs 36.3 (+4.1). Importantly, open ended accuracy improves on every benchmarks while MCQA scores remain competitive. Models trained with ReVeL data achieves a 40.4 overall score, compared to 31.3 for R1-OneVision-7B, 37.2 for Mixed-R1-7B, 37.5 for VL-Rethinker-7B. These results indicate that verifiable OpenQA align better with transferable reasoning and real-world usage, improving both open-ended performance and the combined overall metric.

% \subsection{Performance drop between MCQA and OpenQA}
\subsection{Performance gap in MCQA and OpenQA}
To further quantify the discrepancy in model capabilities between MCQA and OpenQA, we conduct a comparative analysis of model performance in MCQA and OpenQA setting with two rewritten datasets (ViRL and Mixed-R1).

The comprehensive results of this evaluation are presented in Table ~\ref{tab:main-results}. The result reveal a consistent and substantial performance degradation across all evaluated models when transitioning from the MCQA to the OpenQA format,
even strong MLLMs such as GPT-5 and Gemini-2.5 flash are not immune to this effect. For instance, GPT-5's accuracy on the MMMU benchmark drops by 19.8 points (from 79.2\% to 59.5\%), and Gemini-2.5 flash's accuracy on EMMA decreases by 15.7 points. This indicates that the challenge of OpenQA is a fundamental problem that affects even the most advanced models.

And we observe that the performance gap is often more pronounced for open-weight models. For example, R1-OneVision-7B exhibits a staggering 24.2-point drop on EMMA, while InternVL3-8B's performance on MMMU plummets by 27.9 points. This suggests that many open-weight MLLMs may particularly overfit the MCQA format, which is prevalent in many VQA datasets. 

\begin{table*}[h]
  \centering
  \small
  \caption{Overall accuracy (\%). Accuracy drop between MCQA and OpenQA is marked after $\downarrow$. Bold numbers indicate the smallest drop across open-sourced models}
  \setlength{\tabcolsep}{4pt}
  \begin{tabular}{lcccccccc}
    \toprule
    \multicolumn{1}{c}{\multirow{2}{*}{Model}}  & \multicolumn{2}{c}{EMMA} & \multicolumn{2}{c}{MMMU} & \multicolumn{2}{c}{MME-RealWorld} & \multicolumn{2}{c}{MMLU-Pro} \\
    \cmidrule(lr){2-3} \cmidrule(lr){4-5} \cmidrule(lr){6-7} \cmidrule(lr){8-9}
    & MCQA & OpenQA & MCQA & OpenQA & MCQA & OpenQA & MCQA & OpenQA \\
    \midrule
    \multicolumn{9}{c}{\textit{Proprietary Models}} \\
    \midrule
    GPT-5 & 42.0 & 36.0 ($\downarrow$6.0) & 79.2 & 59.5 ($\downarrow$19.8) & 57.8 & 42.4 ($\downarrow$15.4) & 84.6 & 67.6 ($\downarrow$17.0) \\
    GPT-5 mini & 42.8 & 35.0 ($\downarrow$7.8) & 75.2 & 55.5 ($\downarrow$19.7) & 58.3 & 43.7 ($\downarrow$14.6) & 78.7 & 63.8 ($\downarrow$14.9) \\
    GPT-4.1 & 36.4 & 27.3 ($\downarrow$9.1) & 71.7 & 56.1 ($\downarrow$15.5) & 52.7 & 39.6 ($\downarrow$13.1) & 81.2 & 67.1 ($\downarrow$14.1) \\
    GPT-4.1 mini & 40.2 & 22.3 ($\downarrow$17.9) & 65.3 & 51.6 ($\downarrow$13.7) & 54.8 & 44.0 ($\downarrow$10.9) & 75.4 & 64.4 ($\downarrow$11.0) \\
    Gemini-2.5 flash & 49.2 & 33.6 ($\downarrow$15.7) & 69.6 & 57.7 ($\downarrow$11.9) & 57.3 & 46.5 ($\downarrow$10.8) & 78.3 & 63.8 ($\downarrow$14.5) \\
    \midrule
    \multicolumn{9}{c}{\textit{Open-Source Models}} \\
    \midrule
    InternVL3-78B & 34.6 & 20.8 ($\downarrow$13.8) & 67.7 & 51.5 ($\downarrow$\textbf{16.2}) & 48.9 & 31.4 ($\downarrow$17.5) & 70.9 & 57.0 ($\downarrow$13.9) \\
    InternVL3-8B & 32.2 & 14.5 ($\downarrow$17.6) & 60.0 & 32.1 ($\downarrow$27.9) & 49.6 & 33.2 ($\downarrow$16.4) & 55.3 & 39.0($\downarrow$16.3) \\
    Qwen3-VL-8B-Instruct & 42.1 & 23.0 ($\downarrow$19.1) & 68.5 & 46.5 ($\downarrow$22.0) & 51.7 & 41.5 ($\downarrow$10.2) & 74.6 & 60.7 ($\downarrow$13.9) \\
    R1-OneVision-7B & 28.9 & 4.7 ($\downarrow$24.2) & 42.2 & 23.9 ($\downarrow$18.3) & 44.6 & 31.6 ($\downarrow$13.0) & 42.5 & 32.3 ($\downarrow$10.2) \\
    Mixed-R1-7B & 29.8 & 13.2 ($\downarrow$16.7) & 56.3 & 30.6 ($\downarrow$25.8) & 45.6 & 32.8 ($\downarrow$12.8) & 51.4 & 37.7 ($\downarrow$13.7) \\
    VL-Rethinker-7B & 30.6 & 14.9 ($\downarrow$15.8) & 53.9 & 33.4 ($\downarrow$20.5) & 44.3 & 32.7 ($\downarrow$11.6) & 52.4 & 37.6 ($\downarrow$14.8)  \\
    Qwen2.5-VL-72B & 35.9 & 20.6 ($\downarrow$15.3) & 68.2 & 47.9 ($\downarrow$20.3) & 48.4 & 37.4 ($\downarrow$11.0) & 70.8 & 57.6 ($\downarrow$13.2) \\
    \midrule
    Qwen2.5-VL-3B & 27.4 & 5.7 ($\downarrow$21.7) & 44.3 & 23.3 ($\downarrow$21.0) & 35.9 & 26.6 ($\downarrow$9.2) & 38.7 & 29.6 ($\downarrow$9.1) \\
    \quad+OpenQA(ViRL) & 29.8 & 18.6 ($\downarrow$\textbf{11.3}) & 49.4 & 27.4 ($\downarrow$22.0) & 41.2 & 31.9 ($\downarrow$9.3) & 42.2 & 34.1 \textbf{($\downarrow$8.1)}\\
    \quad+OpenQA(Mixed-R1) & 31.4 & 17.2 ($\downarrow$14.1) & 46.3 & 29.8 ($\downarrow$16.5) & 38.0 & 36.3\textbf{ ($\downarrow$1.7)} & 43.3 & 32.8 ($\downarrow$10.5)\\
    Qwen2.5-VL-7B & 28.9 & 10.2 ($\downarrow$18.7) & 51.9 & 31.9 ($\downarrow$20.0) & 44.8 & 32.8 ($\downarrow$12.0) & 49.1 & 39.0 ($\downarrow$10.1) \\
    \quad+OpenQA(ViRL) & 29.2 & 17.1 ($\downarrow$12.1) & 56.4 & 37.0 ($\downarrow$19.5) & 50.6 & 38.8 ($\downarrow$11.7) & 51.1 & 43.0 ($\downarrow$8.1)\\
    \quad+OpenQA(Mixed-R1) & 29.4 & 15.1 ($\downarrow$14.4) & 56.1 & 34.1 ($\downarrow$22.0) & 51.9 & 39.6 ($\downarrow$12.3) & 53.8 & 40.9 ($\downarrow$12.9) \\
    % Qwen3-VL-8B-Thinking & 33.8 & 23.2 ($\downarrow$10.6) & 69.0 & 44.4 ($\downarrow$24.5) & 42.6 & \textbf{38.7 ($\downarrow$3.8)} & 73.4 & 55.4 ($\downarrow$18.0) \\
    \bottomrule
  \end{tabular}
  \label{tab:main-results}
\end{table*}

\section{Related work}
\textbf{Multiple-choice question answering (MCQA)} 
%For many years, students and faculties in education have preferred MCQA as an assessment tool due to its ease of implementation and scoring
has been a popularly used assessment tool for ages due to simplified grading \citep{Simkin2005MultipleChoice, makingsenseMCQ, LinguisticMCQ, Balepur2025FlawsMCQ, alzahrani-etal-2024-MCQSensitivity, SensitivitytoTheOrder}.
% This convenience led to its widespread adoption for constraining the output space when evaluating large (multi-modal) language and vision-language models \citep{Hendrycks2020MMLU, Wang2024MMLUPro, Yue2023MMMU, arcbench, Yue2024MMMUPro, Liu2024IIBench}. 
This convenience led to its wide adoption for evaluation of large language models~\citep{Hendrycks2020MMLU,Wang2024MMLUPro}, and in particular vision-language models \citep{Yue2023MMMU, arcbench, Yue2024MMMUPro, Liu2024IIBench} because of more diverse wording choices in describing many visual concepts or scenes.
%Early evaluation methods for these models often relied on likelihood-based scoring, \citep{alzahrani-etal-2024-MCQSensitivity, Wang2024Myanswerisc, Hu2023PromptingIN}.
However, MCQA has many shortcuts. Performance can drop dramatically simply from changing an option's placement\citep{Zheng2023LMSelectBias, Molfese2025inconMultiChoice}. While mitigation strategies—such as better distractors, more options, randomized order, or 'select all that apply' formats \citep{Zhang2025AutomatedOpen2Choice, yu-etal-2024-KnowledgeEnhancingDistractor, Zheng2023LMSelectBias, zhou-etal-2024-revisiting, Xu2025SATABENCH} reduced some biases. And models typically cannot reject all options when the correct answer is absent \citep{Goral2024IncorrectMCQ, Tam2025NoneOPfTheAbove}. Some recent work has shown that reasoning models are good at exploiting the information in the options, implying the performance may be inflated \citep{Balepur2024ArtifactsOA, Raman2025MCQExploiter}. Recognizing these issues, the community's shift to open-ended evaluation faces its own challenges. Rule-based, short-answer benchmarks \citep{realwordqa, Wang2024CharXiv} are limited in scope, while general open-ended formats rely on an LLM-as-a-judge. Furthermore, simply remove options must discard a significant portion of unsuitable items and still depend on an LLM-Judge for evaluation \citep{Myrzakhan2024OpenLLMLeaderboardFM}. These works analyse the flaws of MCQA but do not try to propose a method to mitigate these shortcuts. These analyses focus on identifying the flaws of MCQA rather than proposing systematic mitigation strategies.

\textbf{Multimodal reinforcement learning}:
Many visual reasoning datasets
%MCQA remains the dominant format in large-scale visual reasoning datasets. 
are predominantly designed in an MCQA format.
For instance, earlier datasets such as ScienceQA \citep{Lu2022ScienceQA}, AI2D \citep{Kembhavi2016AI2D}, Geometry3K \citep{Lu2021InterGPSIG}, and GeoQA-Plus \citep{Chen2021GeoQAAG} are entirely formed by multiple-choice questions. This trend continues in recent MLLMs designed for general-purpose reasoning, such as Mixed-R1 \citep{Xu2025MixedR1}, R1-OneVision \citep{Yang2025R1Onevision}, and VL-Rethinker\citep{Wang2025VLRethinker}, which all employ a considerable proportion of choice-based items, accounting for 43\%, 80\%, and 45\% of their data, respectively. Our work is built on these visual reasoning datasets and explores open-form rewriting from those MCQA samples.
%\section{Limitation and Conclusion}

\section{Limitations}
We acknowledge several limitations in our proposed pipeline. First, the rewriting and classification phases, while highly accurate, are not perfect and may occasionally introduce errors. Hopefully such errors could diminish when the LLM components are getting stronger and stronger in the future. Second, our work focuses on converting the format of evaluation to be more robust and efficient, without addressing the inherent fallibility of the LLM-judge itself. Issues such as positional bias, verbosity bias, or factual inaccuracies within the LLM-judge \citep{Chen2024MLLMasaJudge} are orthogonal to our contribution. We deliberately sidestep some of these known issues; for instance, questions in the EMMA dataset requiring the validation of SMILES chemical structures were intentionally converted to a Per-Option Verification format. This leverages rule-based checking and avoids relying on an LLM-judge for a domain-specific task, thereby mitigating a potential failure point of LLM-based evaluation. There are several directions for future research. One key avenue is to extend our framework beyond QA to other NLP tasks, such as long-form generation, where evaluation remains a major challenge. Finally, developing adaptive evaluation systems that can dynamically choose the most appropriate and cost effective judging mechanism based on the question's complexity and the model's response would be a valuable next step.

\section{Conclusions} 
In this work, we systematically demonstrated the fragility of MCQA format for both evaluation and reinforcement fine-tuning. We found that MCQA metrics significantly overestimate model capabilities, and RFT on MCQA data reinforces format-specific shortcuts, harming open-ended generalization. To solve this, we propose ReVeL, a framework that rewrite MCQA into verifiable OpenQA by categorizing questions for a hybird evaluation scheme. Applying ReVeL to RFT, we found that models trained on our rewritten OpenQA data achieved approximately a 6-point improvement in open-ended accuracy while maintaining performance on original MCQA benchmarks, confirming its role in fostering more robust and transferable reasoning.

{
    \small
    \bibliographystyle{ieeenat_fullname}
    \bibliography{main}

@String(CVPR= {IEEE Conf. Comput. Vis. Pattern Recog.})

@String(CVPR  = {CVPR})

@article{Hendrycks2020MMLU,
  title={Measuring Massive Multitask Language Understanding},
  author={Dan Hendrycks and Collin Burns and Steven Basart and Andy Zou and Mantas Mazeika and Dawn Xiaodong Song and Jacob Steinhardt},
  journal={ArXiv},
  year={2020},
  volume={abs/2009.03300},
  url={https://api.semanticscholar.org/CorpusID:221516475}
}

@article{Yue2023MMMU,
  title={MMMU: A Massive Multi-Discipline Multimodal Understanding and Reasoning Benchmark for Expert AGI},
  author={Xiang Yue and Yuansheng Ni and Kai Zhang and Tianyu Zheng and Ruoqi Liu and Ge Zhang and Samuel Stevens and Dongfu Jiang and Weiming Ren and Yuxuan Sun and Cong Wei and Botao Yu and Ruibin Yuan and Renliang Sun and Ming Yin and Boyuan Zheng and Zhenzhu Yang and Yibo Liu and Wenhao Huang and Huan Sun and Yu Su and Wenhu Chen},
  journal={2024 IEEE/CVF Conference on Computer Vision and Pattern Recognition (CVPR)},
  year={2023},
  pages={9556-9567},
  url={https://api.semanticscholar.org/CorpusID:265466525}
}

@article{Liu2023MMBench,
  title={MMBench: Is Your Multi-modal Model an All-around Player?},
  author={Yuanzhan Liu and Haodong Duan and Yuanhan Zhang and Bo Li and Songyang Zhang and Wangbo Zhao and Yike Yuan and Jiaqi Wang and Conghui He and Ziwei Liu and Kai Chen and Dahua Lin},
  journal={ArXiv},
  year={2023},
  volume={abs/2307.06281},
  url={https://api.semanticscholar.org/CorpusID:259837088}
}

@article{Kembhavi2016AI2D,
  title={A Diagram is Worth a Dozen Images},
  author={Aniruddha Kembhavi and Michael Salvato and Eric Kolve and Minjoon Seo and Hannaneh Hajishirzi and Ali Farhadi},
  journal={ArXiv},
  year={2016},
  volume={abs/1603.07396},
  url={https://api.semanticscholar.org/CorpusID:2682274}
}

@article{Wang2024MMLUPro,
  title={MMLU-Pro: A More Robust and Challenging Multi-Task Language Understanding Benchmark},
  author={Yubo Wang and Xueguang Ma and Ge Zhang and Yuansheng Ni and Abhranil Chandra and Shiguang Guo and Weiming Ren and Aaran Arulraj and Xuan He and Ziyan Jiang and Tianle Li and Max W.F. Ku and Kai Wang and Alex Zhuang and Rongqi "Richard" Fan and Xiang Yue and Wenhu Chen},
  journal={ArXiv},
  year={2024},
  volume={abs/2406.01574},
  url={https://api.semanticscholar.org/CorpusID:270210486}
}

@article{Zheng2023LMSelectBias,
  title={Large Language Models Are Not Robust Multiple Choice Selectors},
  author={Chujie Zheng and Hao Zhou and Fandong Meng and Jie Zhou and Minlie Huang},
  journal={ArXiv},
  year={2023},
  volume={abs/2309.03882},
  url={https://api.semanticscholar.org/CorpusID:261582594}
}

@InProceedings{Zhang2025AutomatedOpen2Choice,
    author    = {Zhang, Yuhui and Su, Yuchang and Liu, Yiming and Wang, Xiaohan and Burgess, James and Sui, Elaine and Wang, Chenyu and Aklilu, Josiah and Lozano, Alejandro and Wei, Anjiang and Schmidt, Ludwig and Yeung-Levy, Serena},
    title     = {Automated Generation of Challenging Multiple-Choice Questions for Vision Language Model Evaluation},
    booktitle = {Proceedings of the Computer Vision and Pattern Recognition Conference (CVPR)},
    month     = {June},
    year      = {2025},
    pages     = {29580-29590}
}

@article{Yue2024MMMUPro,
  title={MMMU-Pro: A More Robust Multi-discipline Multimodal Understanding Benchmark},
  author={Xiang Yue and Tianyu Zheng and Yuansheng Ni and Yubo Wang and Kai Zhang and Shengbang Tong and Yuxuan Sun and Ming Yin and Botao Yu and Ge Zhang and Huan Sun and Yu Su and Wenhu Chen and Graham Neubig},
  journal={ArXiv},
  year={2024},
  volume={abs/2409.02813},
  url={https://api.semanticscholar.org/CorpusID:272397682}
}

@inproceedings{yu-etal-2024-KnowledgeEnhancingDistractor,
    title = "Enhancing Distractor Generation for Multiple-Choice Questions with Retrieval Augmented Pretraining and Knowledge Graph Integration",
    author = "Yu, Han Cheng  and
      Shih, Yu An  and
      Law, Kin Man  and
      Hsieh, KaiYu  and
      Cheng, Yu Chen  and
      Ho, Hsin Chih  and
      Lin, Zih An  and
      Hsu, Wen-Chuan  and
      Fan, Yao-Chung",
    editor = "Ku, Lun-Wei  and
      Martins, Andre  and
      Srikumar, Vivek",
    booktitle = "Findings of the Association for Computational Linguistics: ACL 2024",
    month = aug,
    year = "2024",
    address = "Bangkok, Thailand",
    publisher = "Association for Computational Linguistics",
    url = "https://aclanthology.org/2024.findings-acl.655/",
    doi = "10.18653/v1/2024.findings-acl.655",
    pages = "11019--11029"
}

@misc{realwordqa,
  author = {xAI},
  title = {Introducing Grok-1.5V and RealWorldQA Benchmark},
  year = {2024},
  url = {https://x.ai/news/grok-1.5v}
}

@article{openai2023gpt4v,
  title={GPT-4V(ision) System Card},
  author={OpenAI},
  journal={OpenAI Research},
  year={2023},
  url = "https://cdn.openai.com/papers/GPTV_System_Card.pdf",
}

@article{Bai2025Qwen25VLTR,
  title={Qwen2. 5-vl technical report},
  author={Bai, Shuai and Chen, Keqin and Liu, Xuejing and Wang, Jialin and Ge, Wenbin and Song, Sibo and Dang, Kai and Wang, Peng and Wang, Shijie and Tang, Jun and others},
  journal={arXiv preprint arXiv:2502.13923},
  year={2025}
}

@misc{Anthropic2024Claude4,
  author       = {Anthropic},
  title        = {Claude 4},
  howpublished = {\url{https://www.anthropic.com/news/claude-4}},
  year         = {2025},
  note         = {Accessed: 2025-07-11}
}

@misc{Google2025Gemini25pro,
  author       = {Google},
  title        = {Gemini2.5 pro},
  howpublished = {\url{https://deepmind.google/models/gemini/}},
  year         = {2025},
  note         = {Accessed: 2025-07-11}
}

@inproceedings{alzahrani-etal-2024-MCQSensitivity,
    title = "When Benchmarks are Targets: Revealing the Sensitivity of Large Language Model Leaderboards",
    author = "Alzahrani, Norah  and
      Alyahya, Hisham  and
      Alnumay, Yazeed  and
      AlRashed, Sultan  and
      Alsubaie, Shaykhah  and
      Almushayqih, Yousef  and
      Mirza, Faisal  and
      Alotaibi, Nouf  and
      Al-Twairesh, Nora  and
      Alowisheq, Areeb  and
      Bari, M Saiful  and
      Khan, Haidar",
    editor = "Ku, Lun-Wei  and
      Martins, Andre  and
      Srikumar, Vivek",
    booktitle = "Proceedings of the 62nd Annual Meeting of the Association for Computational Linguistics (Volume 1: Long Papers)",
    month = aug,
    year = "2024",
    address = "Bangkok, Thailand",
    publisher = "Association for Computational Linguistics",
    url = "https://aclanthology.org/2024.acl-long.744/",
    doi = "10.18653/v1/2024.acl-long.744",
    pages = "13787--13805"
}

@article{Molfese2025inconMultiChoice,
  title={Right Answer, Wrong Score: Uncovering the Inconsistencies of LLM Evaluation in Multiple-Choice Question Answering},
  author={Francesco Maria Molfese and Luca Moroni and Luca Gioffre and Alessandro Scir{\'e} and Simone Conia and Roberto Navigli},
  journal={ArXiv},
  year={2025},
  volume={abs/2503.14996},
  url={https://api.semanticscholar.org/CorpusID:277113217}
}

@article{Tam2025NoneOPfTheAbove,
  title={None of the Above, Less of the Right: Parallel Patterns between Humans and LLMs on Multi-Choice Questions Answering},
  author={Zhi Rui Tam and Cheng-Kuang Wu and Chieh-Yen Lin and Yun-Nung Chen},
  journal={ArXiv},
  year={2025},
  volume={abs/2503.01550},
  url={https://api.semanticscholar.org/CorpusID:276768098}
}

@article{Simpleqa,
  title={Measuring short-form factuality in large language models},
  author={Jason Wei and Nguyen Karina and Hyung Won Chung and Yunxin Joy Jiao and Spencer Papay and Amelia Glaese and John Schulman and William Fedus},
  journal={ArXiv},
  year={2024},
  volume={abs/2411.04368},
  url={https://api.semanticscholar.org/CorpusID:273877483}
}

@inproceedings{Goral2024IncorrectMCQ,
  title={Wait, that's not an option: LLMs Robustness with Incorrect Multiple-Choice Options},
  author={Gracjan G'oral and Emilia Wiśnios and Piotr Sankowski and Pawel Budzianowski},
  year={2024},
  url={https://api.semanticscholar.org/CorpusID:272367487}
}

@article{Simkin2005MultipleChoice,
  title={Multiple-Choice Tests and Student Understanding: What Is the Connection?},
  author={Mark G. Simkin and William L. Kuechler},
  journal={Decision Sciences Journal of Innovative Education},
  year={2005},
  volume={3},
  pages={73-98},
  url={https://api.semanticscholar.org/CorpusID:6239312}
}

@article{LinguisticMCQ,
author = {Paxton, Moragh},
year = {2000},
month = {06},
pages = {109-119},
title = {A Linguistic Perspective on Multiple Choice Questioning},
volume = {25},
journal = {Assessment \& Evaluation in Higher Education - ASSESS EVAL HIGH EDUC},
doi = {10.1080/713611429}
}

@article{makingsenseMCQ,
author = {Dufresne, Robert and Leonard, William and Gerace, William},
year = {2002},
month = {03},
pages = {174-180},
title = {Making sense of students' answers to multiple-choice questions},
volume = {40},
journal = {The Physics Teacher},
doi = {10.1119/1.1466554}
}

@inproceedings{Balepur2025FlawsMCQ,
  title={Which of These Best Describes Multiple Choice Evaluation with LLMs? A) Forced B) Flawed C) Fixable D) All of the Above},
  author={Nishant Balepur and Rachel Rudinger and Jordan L. Boyd-Graber},
  booktitle={Annual Meeting of the Association for Computational Linguistics},
  year={2025},
  url={https://api.semanticscholar.org/CorpusID:276482640}
}

@article{Myrzakhan2024OpenLLMLeaderboardFM,
  title={Open-LLM-Leaderboard: From Multi-choice to Open-style Questions for LLMs Evaluation, Benchmark, and Arena},
  author={Aidar Myrzakhan and S. Mahmoud Bsharat and Zhiqiang Shen},
  journal={ArXiv},
  year={2024},
  volume={abs/2406.07545},
  url={https://api.semanticscholar.org/CorpusID:270379906}
}

@misc{Chen2025InternVL2.5,
      title={Expanding Performance Boundaries of Open-Source Multimodal Models with Model, Data, and Test-Time Scaling}, 
      author={Zhe Chen and Weiyun Wang and Yue Cao and Yangzhou Liu and Zhangwei Gao and Erfei Cui and Jinguo Zhu and Shenglong Ye and Hao Tian and Zhaoyang Liu and Lixin Gu and Xuehui Wang and Qingyun Li and Yimin Ren and Zixuan Chen and Jiapeng Luo and Jiahao Wang and Tan Jiang and Bo Wang and Conghui He and Botian Shi and Xingcheng Zhang and Han Lv and Yi Wang and Wenqi Shao and Pei Chu and Zhongying Tu and Tong He and Zhiyong Wu and Huipeng Deng and Jiaye Ge and Kai Chen and Kaipeng Zhang and Limin Wang and Min Dou and Lewei Lu and Xizhou Zhu and Tong Lu and Dahua Lin and Yu Qiao and Jifeng Dai and Wenhai Wang},
      year={2025},
      eprint={2412.05271},
      archivePrefix={arXiv},
      primaryClass={cs.CV},
      url={https://arxiv.org/abs/2412.05271}, 
}

@misc{GPT-5,
    title = {Introducing {GPT}‑5},
    url = {https://openai.com/index/introducing-gpt-5/},
    author = {{OpenAI}},
    month = {August},
    year = {2025}
}

@inproceedings{Moore2023AssessingMCQ,
  title={Assessing the Quality of Multiple-Choice Questions Using GPT-4 and Rule-Based Methods},
  author={Steven Moore and Huy Anh Nguyen and Tianying Chen and John C. Stamper},
  booktitle={European Conference on Technology Enhanced Learning},
  year={2023},
  url={https://api.semanticscholar.org/CorpusID:259937876}
}

@inproceedings{SensitivitytoTheOrder,
  title={Large Language Models Sensitivity to The Order of Options in Multiple-Choice Questions},
  author={Pouya Pezeshkpour and Estevam Hruschka},
  booktitle={NAACL-HLT},
  year={2023},
  url={https://api.semanticscholar.org/CorpusID:261064970}
}

@article{arcbench,
  title={Think you have Solved Question Answering? Try ARC, the AI2 Reasoning Challenge},
  author={Peter Clark and Isaac Cowhey and Oren Etzioni and Tushar Khot and Ashish Sabharwal and Carissa Schoenick and Oyvind Tafjord},
  journal={ArXiv},
  year={2018},
  volume={abs/1803.05457},
  url={https://api.semanticscholar.org/CorpusID:3922816}
}

@article{Liu2024IIBench,
  title={II-Bench: An Image Implication Understanding Benchmark for Multimodal Large Language Models},
  author={Ziqiang Liu and Feiteng Fang and Xi Feng and Xinrun Du and Chenhao Zhang and Zekun Moore Wang and Yuelin Bai and Qixuan Zhao and Liyang Fan and Chengguang Gan and Hongquan Lin and Jiaming Li and Yuansheng Ni and Haihong Wu and Yaswanth Narsupalli and Zhigang Zheng and Chengming Li and Xiping Hu and Ruifeng Xu and Xiaojun Chen and Min Yang and Jiaheng Liu and Ruibo Liu and Wenhao Huang and Ge Zhang and Shiwen Ni},
  journal={ArXiv},
  year={2024},
  volume={abs/2406.05862},
  url={https://api.semanticscholar.org/CorpusID:270371872}
}

@article{Wang2024CharXiv,
  title={CharXiv: Charting Gaps in Realistic Chart Understanding in Multimodal LLMs},
  author={Zirui Wang and Mengzhou Xia and Luxi He and Howard Chen and Yitao Liu and Richard Zhu and Kaiqu Liang and Xindi Wu and Haotian Liu and Sadhika Malladi and Alexis Chevalier and Sanjeev Arora and Danqi Chen},
  journal={ArXiv},
  year={2024},
  volume={abs/2406.18521},
  url={https://api.semanticscholar.org/CorpusID:270737638}
}

@article{Hao2025EMMA,
  title={Can MLLMs Reason in Multimodality? EMMA: An Enhanced MultiModal ReAsoning Benchmark},
  author={Yunzhuo Hao and Jiawei Gu and Huichen Will Wang and Linjie Li and Zhengyuan Yang and Lijuan Wang and Yu Cheng},
  journal={ArXiv},
  year={2025},
  volume={abs/2501.05444},
  url={https://api.semanticscholar.org/CorpusID:275405458}
}

@article{Zhang2024MMERealWorld,
  title={MME-RealWorld: Could Your Multimodal LLM Challenge High-Resolution Real-World Scenarios that are Difficult for Humans?},
  author={Yi-Fan Zhang and Huanyu Zhang and Haochen Tian and Chaoyou Fu and Shuangqing Zhang and Jun Wu and Feng Li and Kun Wang and Qingsong Wen and Zhang Zhang and Liang Wang and Rong Jin and Tien-Ping Tan},
  journal={ArXiv},
  year={2024},
  volume={abs/2408.13257},
  url={https://api.semanticscholar.org/CorpusID:271947320}
}

@article{Raman2025MCQExploiter,
  title={Reasoning Models are Test Exploiters: Rethinking Multiple-Choice},
  author={Narun K. Raman and Taylor Lundy and Kevin Leyton-Brown},
  journal={ArXiv},
  year={2025},
  volume={abs/2507.15337},
  url={https://api.semanticscholar.org/CorpusID:280671880}
}

@inproceedings{Chen2024MLLMasaJudge,
  title={MLLM-as-a-Judge: Assessing Multimodal LLM-as-a-Judge with Vision-Language Benchmark},
  author={Dongping Chen and Ruoxi Chen and Shilin Zhang and Yinuo Liu and Yaochen Wang and Huichi Zhou and Qihui Zhang and Pan Zhou and Yao Wan and Lichao Sun},
  booktitle={International Conference on Machine Learning},
  year={2024},
  url={https://api.semanticscholar.org/CorpusID:267523079}
}

@article{Yang2025R1Onevision,
  title={R1-Onevision: Advancing Generalized Multimodal Reasoning through Cross-Modal Formalization},
  author={Yi Yang and Xiaoxuan He and Hongkun Pan and Xiyan Jiang and Yan Deng and Xingtao Yang and Haoyu Lu and Dacheng Yin and Fengyun Rao and Minfeng Zhu and Bo Zhang and Wei Chen},
  journal={ArXiv},
  year={2025},
  volume={abs/2503.10615},
  url={https://api.semanticscholar.org/CorpusID:276961560}
}

@article{Wang2025VLRethinker,
  title={VL-Rethinker: Incentivizing Self-Reflection of Vision-Language Models with Reinforcement Learning},
  author={Haozhe Wang and Chao Qu and Zuming Huang and Wei Chu and Fangzhen Lin and Wenhu Chen},
  journal={ArXiv},
  year={2025},
  volume={abs/2504.08837},
  url={https://api.semanticscholar.org/CorpusID:277781277}
}

@article{Lu2022ScienceQA,
  title={Learn to Explain: Multimodal Reasoning via Thought Chains for Science Question Answering},
  author={Pan Lu and Swaroop Mishra and Tony Xia and Liang Qiu and Kai-Wei Chang and Song-Chun Zhu and Oyvind Tafjord and Peter Clark and A. Kalyan},
  journal={ArXiv},
  year={2022},
  volume={abs/2209.09513},
  url={https://api.semanticscholar.org/CorpusID:252383606}
}

@article{Xu2025MixedR1,
  title={Mixed-R1: Unified Reward Perspective For Reasoning Capability in Multimodal Large Language Models},
  author={Shilin Xu and Yanwei Li and Rui Yang and Tao Zhang and Yueyi Sun and Wei Chow and Linfeng Li and Hang Song and Qi Xu and Yunhai Tong and Xiangtai Li and Hao Fei},
  journal={ArXiv},
  year={2025},
  volume={abs/2505.24164},
  url={https://api.semanticscholar.org/CorpusID:279070666}
}

@article{Lyu2024BeyondPro,
  title={Beyond Probabilities: Unveiling the Misalignment in Evaluating Large Language Models},
  author={Chenyang Lyu and Minghao Wu and Alham Fikri Aji},
  journal={ArXiv},
  year={2024},
  volume={abs/2402.13887},
  url={https://api.semanticscholar.org/CorpusID:267770168}
}

@inproceedings{Balepur2024ArtifactsOA,
  title={Artifacts or Abduction: How Do LLMs Answer Multiple-Choice Questions Without the Question?},
  author={Nishant Balepur and Abhilasha Ravichander and Rachel Rudinger},
  booktitle={Annual Meeting of the Association for Computational Linguistics},
  year={2024},
  url={https://api.semanticscholar.org/CorpusID:267760229}
}

@article{Xu2025SATABENCH,
  title={SATA-BENCH: Select All That Apply Benchmark for Multiple Choice Questions},
  author={Weijie Xu and Shixian Cui and Xi Fang and Chi Xue and Stephanie Eckman and Chandan K. Reddy},
  journal={ArXiv},
  year={2025},
  volume={abs/2506.00643},
  url={https://api.semanticscholar.org/CorpusID:279074686}
}

@inproceedings{zhou-etal-2024-revisiting,
    title = "Revisiting the Self-Consistency Challenges in Multi-Choice Question Formats for Large Language Model Evaluation",
    author = "Zhou, Wenjie  and
      Wang, Qiang  and
      Xu, Mingzhou  and
      Chen, Ming  and
      Duan, Xiangyu",
    editor = "Calzolari, Nicoletta  and
      Kan, Min-Yen  and
      Hoste, Veronique  and
      Lenci, Alessandro  and
      Sakti, Sakriani  and
      Xue, Nianwen",
    booktitle = "Proceedings of the 2024 Joint International Conference on Computational Linguistics, Language Resources and Evaluation (LREC-COLING 2024)",
    month = may,
    year = "2024",
    address = "Torino, Italia",
    publisher = "ELRA and ICCL",
    url = "https://aclanthology.org/2024.lrec-main.1229/",
    pages = "14103--14110"
}

@article{Wang2025VisualSimpleQA,
  title={VisualSimpleQA: A Benchmark for Decoupled Evaluation of Large Vision-Language Models in Fact-Seeking Question Answering},
  author={Yanling Wang and Yihan Zhao and Xiaodong Chen and Shasha Guo and Lixin Liu and Haoyang Li and Yong Xiao and Jing Zhang and Qi Li and Ke Xu},
  journal={ArXiv},
  year={2025},
  volume={abs/2503.06492},
  url={https://api.semanticscholar.org/CorpusID:276903700}
}

@inproceedings{Pei2025EgoThinker,
  title={EgoThinker: Unveiling Egocentric Reasoning with Spatio-Temporal CoT},
  author={Baoqi Pei and Yifei Huang and Jilan Xu and Yuping He and Guo Chen and Fei Wu and Yu Qiao and Jiangmiao Pang},
  year={2025},
  url={https://api.semanticscholar.org/CorpusID:282389525}
}

@inproceedings{Lu2021InterGPSIG,
  title={Inter-GPS: Interpretable Geometry Problem Solving with Formal Language and Symbolic Reasoning},
  author={Pan Lu and Ran Gong and Shibiao Jiang and Liang Qiu and Siyuan Huang and Xiaodan Liang and Song-Chun Zhu},
  booktitle={Annual Meeting of the Association for Computational Linguistics},
  year={2021},
  url={https://api.semanticscholar.org/CorpusID:234337054}
}

@article{Chen2021GeoQAAG,
  title={GeoQA: A Geometric Question Answering Benchmark Towards Multimodal Numerical Reasoning},
  author={Jiaqi Chen and Jianheng Tang and Jinghui Qin and Xiaodan Liang and Lingbo Liu and Eric P. Xing and Liang Lin},
  journal={ArXiv},
  year={2021},
  volume={abs/2105.14517},
  url={https://api.semanticscholar.org/CorpusID:235253782}
}

@article{Shao2024DeepSeekMath,
  title={DeepSeekMath: Pushing the Limits of Mathematical Reasoning in Open Language Models},
  author={Zhihong Shao and Peiyi Wang and Qihao Zhu and Runxin Xu and Jun-Mei Song and Mingchuan Zhang and Y. K. Li and Yu Wu and Daya Guo},
  journal={ArXiv},
  year={2024},
  volume={abs/2402.03300},
  url={https://api.semanticscholar.org/CorpusID:267412607}
}
}
\appendix
% \section{Appendix}
% \input{iclr2026/examples/simpleqa}
\clearpage
\setcounter{page}{1}
% \appendix
\section{Details of Removing Options From MCQA}

\tcbset{
  aibox/.style={
    width=\linewidth,
    top=8pt,
    bottom=4pt,
    colback=blue!6!white,
    colframe=black,
    colbacktitle=black,
    % enhanced,
    center,
    % attach boxed title to top left={yshift=-0.1in,xshift=0.15in},
    % boxed title style={boxrule=0pt,colframe=white,},
  }
}
\newtcolorbox{AIbox}[2][]{aibox,title=#2,#1}
\label{sec:mcq2open}
\subsection{A Filtering Pipeline for Self-Sufficient Questions}

\begin{figure*}[htbp]
\centering
\includegraphics[width=0.7\linewidth]{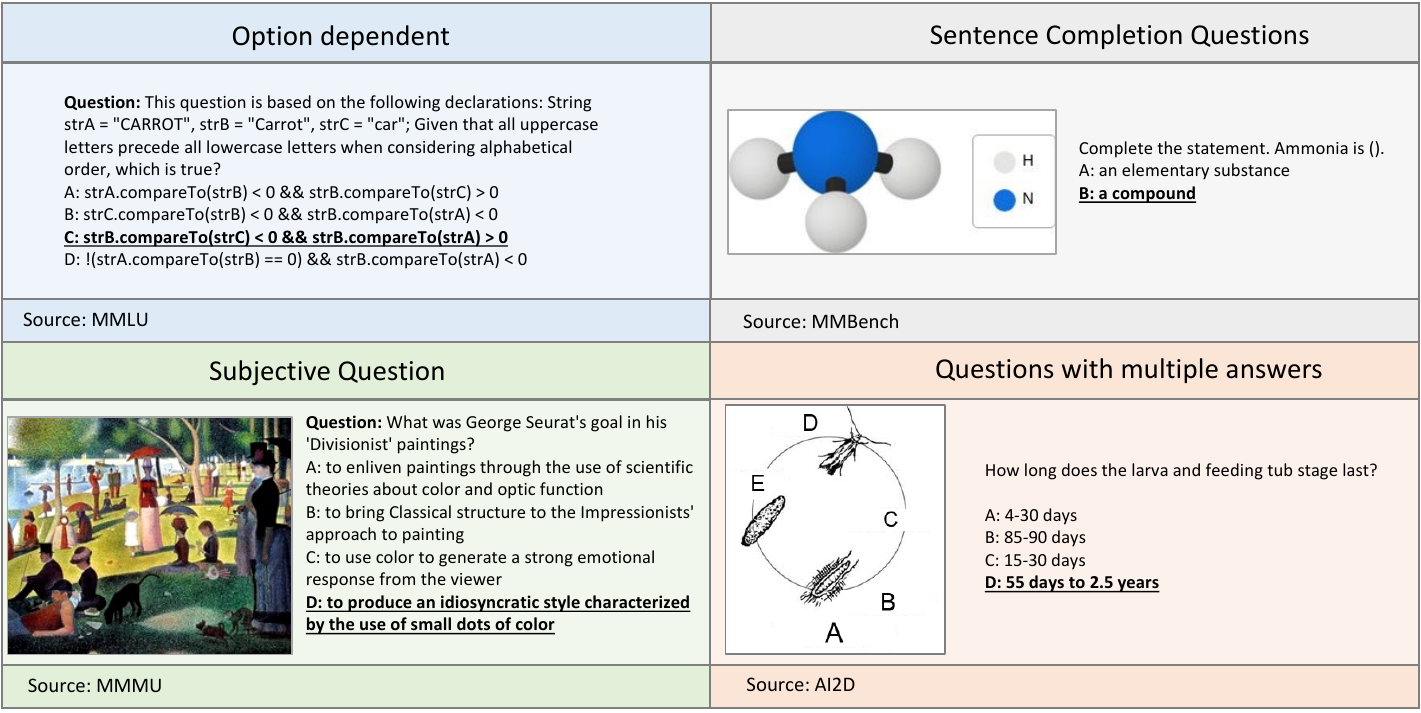}
\caption{Common characteristics that make a multiple-choice question unsuitable for direct conversion to a free-form format.}
\label{fig:types}
\end{figure*}

To create a dataset for our option-free evaluation, we developed a two-stage pipeline to systematically filter existing benchmarks and retain only self-sufficient questions suitable for a generative format.

\textbf{Stage 1: Question Validity Filtering.} 
We first exclude questions that are fundamentally unsuitable for free-form conversion. This stage combines heuristic rules (e.g., removing questions with long, paragraph-style answers likely to be subjective) with a prompted LLM that identifies and removes questions exhibiting option dependency, subjectivity, or underspecification.

\textbf{Stage 2: Answer Uniqueness Verification.} Questions that pass the first stage are then checked for answer uniqueness. We use an LLM to determine if a question, in the absence of options, could yield multiple substantively different but equally valid answers. Only questions with a single, unambiguous correct answer are retained. The filtering prompts used to guide the LLM in each stage are detailed in Appendix.

We validated this pipeline by annotating a random sample of its outputs. The results, detailed in Table \ref{tab:dataset_stats}, confirm the high fidelity of our method. The low overall False Positive (FP) rate of 2.5\% and False Negative (FN) rate of 3.0\% ensure the soundness of our subsequent experiments.

\begin{table}[h]
\centering
\caption{Statistical of the filtering pipeline. The table detailsthe False Positive (FP) and False Negative (FN) rates across sampled subset (right). FP and FN rates are computed from a sampled subset.}
\label{tab:dataset_stats}
  \centering
  \begin{tabular}{l|c|c}
    \toprule
    Dataset & FP (\%) & FN (\%) \\
    \midrule
    MMLU-Pro & 4.0 & 4.0 \\
    MMMU & 2.0 & 2.0 \\
    \bottomrule
  \end{tabular}
\end{table}

Figure \ref{fig:types} shows 4 primary patterns: \textbf{Option-Dependent Questions:} that explicitly refers to the choices (e.g., "Which of the following..."). \textbf{Sentence Completion Questions:} that is a grammatically incomplete without options (e.g., "The letter E in the diagram represents..."). \textbf{Subjective Questions:} that solicits an opinion rather than an objective, factual answer. \textbf{Questions with Multiple Answers:} where the MCQ format artificially presents only one correct choice.

\section{Failure modes of MCQA}

We conduct an analysis of the underlying causes for the unreliability of MCQA, identifying critical failure modes like reasoning-choice mismatch, positional memorization and option-anchoring.

\subsection{Analysis of Failure Modes: Reasoning-Choice Mismatch}

To pinpoint the cause, we focused specifically on the subset of questions that a model answered correctly in the Standard MCQA setting but incorrectly in the NOTA setting. Within this set of failures, we discovered a frequent reasoning-choice mismatch. As shown in Table \ref{tab:mismatch_results}, models often correctly inference the answer in their reasoning steps. However, upon finding this answer absent from the choices, they fail to select the logical NOTA option. Instead, they revert to strategies like string matching to select an incorrect distractor. The rate of this mismatch is dramatically higher in the NOTA setting compared to the standard MCQA setting, exposing a critical flaw in how models interact with provided options when the correct answer is missing.

\begin{table}[h]
\centering
\begin{tabular}{l|l|c|c}
\toprule
Model & Dataset & MCQA & NOTA  \\
\midrule
InternVL3-78B & MMMU & 17.6\% & 54.1\% \\
\midrule
InternVL3-8B & MMMU & 18.6\% & 58.8\% \\
\midrule
Qwen2.5-VL-72B & MMMU & 12.0\% & 46.9\% \\
\midrule
Qwen2.5-VL-7B& MMMU & 21.9\% & 63.8\% \\
\bottomrule
\end{tabular}
\caption{Reasoning-Choice Mismatch Rates by Model and Dataset}
\label{tab:mismatch_results}
\end{table}

\subsection{Analysis of Failure Modes: Positional Memorization}
Another significant failure mode identified in the NOTA setting is positional memorization. We observed that models often select the same option letter in the NOTA task that corresponded to the correct answer in the original MCQA task, even though the content of that option is now an incorrect distractor. This behavior, quantified in Table \ref{tab:positional_memorization} and Table \ref{tab:mmlu_positional_memorization}, indicates that models develop a shallow heuristic of memorizing answer positions instead of semantically evaluating the options provided. This reliance on positional cues rather than content undermines the validity of the evaluation.

\begin{table}[h!]
\centering
\caption{Positional Memorization Statistics by Model}
\label{tab:positional_memorization}
\begin{tabular}{lc}
\toprule
Model &  MMMU \\
\midrule
InternVL3-78B &  30.6\% \\
InternVL3-8B & 36.4\% \\
Qwen2.5-VL-72B  & 42.2\% \\
Qwen2.5-VL-7B  & 40.0\% \\
gemma-3-27b-it  & 35.8\% \\
gpt-4.1   & 42.5\% \\
gpt-4.1-mini  & 33.8\% \\
gpt-5-mini & 33.9\% \\
\bottomrule
\end{tabular}
\end{table}

\begin{table}[h!]
\centering
\caption{MMLU-Pro Positional Memorization Statistics}
\label{tab:mmlu_positional_memorization}
\begin{tabular}{lc}
\toprule
Model & MMLU-Pro \\
\midrule
Meta-Llama-3.3-70B & 28.5\% \\
Qwen2.5-72B & 36.2\% \\
Qwen2.5-7B & 35.6\% \\
\bottomrule
\end{tabular}
\end{table}

\subsection{Example of Option-Anchoring}
\begin{figure*}[ht]
    \centering
    \includegraphics[width=0.95\linewidth]{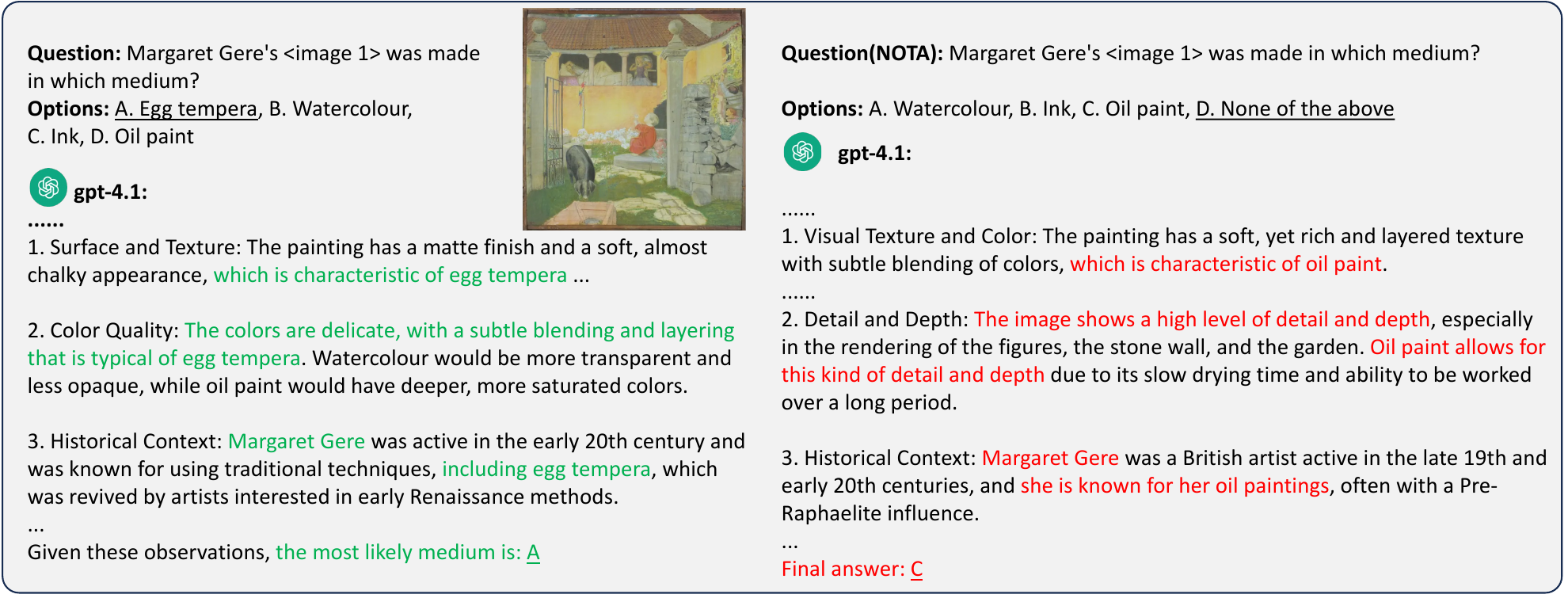}
    \label{fig:option_anchor}
    \caption{Illustration of the Option-Anchoring Phenomenon. \textbf{Left} (Standard MCQA): when "Egg tempera" is an available option, the AI model analyzes the painting's features—such as its matte finish and delicate colors—and concludes they are characteristic of egg tempera.
    \textbf{Right} (NOTA setting): the same model is presented with the same painting, but the "Egg tempera" option is removed and replaced with "None of the above options are correct". The model's reasoning now shifts, describing the painting's texture and detail as characteristic of oil paint.}
\end{figure*}

Option-Anchoring describes how a model's reasoning process becomes tethered to the provided choices. Rather than deriving an answer objectively from the source context (e.g., an image or text), the model first identifies a plausible option and then constructs a post-hoc justification for it. As illustrated in the case study in Figure \ref{fig:option_anchor}, the model's explanation is not a genuine deduction but a rationalization "anchored" to a pre-selected choice. This behavior suggests that the model is not performing true reasoning but rather a form of constrained validation.

\subsection{Deep in Fragility of MCQA: Category Level Effects}
We conduct a category-level analysis of MCQA-NOTA and OpenQA on the filtered items. The effect is not uniform: subjects such as Optical Character Recognition (OCR), Object Localization, and Abstract Algebra consistently exhibit the largest degradations under MCQA--NOTA.

A plausible driver is the \emph{semantic sparsity} of option sets in these domains. For instance, many Object Localization questions present purely numeric options (e.g., A: 3, B: 4, C: 5, D: 6) with minimal contextual content. When a model’s internal reasoning yields an answer not present among the options (e.g., “7”), there are few semantic cues to eliminate the remaining distractors; once the correct option is replaced by NOTA, the model is especially prone to confuse itself and select a distractor rather than NOTA. In contrast, subjects whose options carry richer semantics (full phrases/sentences) provide more opportunities for elimination-by-meaning and show smaller NOTA-induced drops. 

\section{Badcase of HyReV}
While our hybrid pipeline achieves a low overall error rate (~2\%), a closer analysis of the misjudged cases reveals an inherent challenge in rule-based evaluation: the ambiguity of symbolic representation. The primary source of errors is not the pipeline, but rather the vast, often inexhaustible, variations in how a concept can be expressed.

For example, an answer representing a numerical range, such as "1.30 ~ 40.45", presents a significant challenge for any keyword-based or rule-based system. The tilde symbol (~) can be represented in numerous ways in a model's free-form response, including textually ("1.30 to 40.45", "between 1.30 and 40.45"), with different symbols ("1.30 - 40.45"), or in specific formats like LaTeX (1.30 $\sim$ 40.45).

It is computationally infeasible for a rule-based system to enumerate every possible permutation of such representations. Our pipeline is designed to handle common cases, but these edge cases with high representational variance account for the small residual error rate. This is not a flaw in the pipeline's logic but rather a fundamental limitation of deterministic matching when faced with the creative and diverse outputs of modern language models.
\section{Model Behavior after MCQ-to-TF Reformulation}
We analyze model behavior when multiple-choice questions (MCQs) are reformulated into true-false (TF) statements, a transformation that helps mitigate option elimination in MCQA. The Emma dataset focuses on professional physics and chemistry, whereas MMMU-Pro emphasizes high-school and college-level linguistic understanding. Our analysis thus focuses on MMMU-Pro, where semantic reasoning is more central to the observed label imbalance. As shown in Table \ref{table:tf_comparision}, models systematically over-assign true labels compared to the ground-truth annotations after reformulation. 
We define the over-true ratio as:
(Number of answers with >1 correct option) / (Total incorrect answers)
In the MCQ format, models tend to perform comparative reasoning and elimination. When reformulated as independent TF statements, this structure disappears. Without these inter-option cues, models evaluate each statement in isolation and display a stronger bias toward affirmative (``true'') judgments.

\begin{table}[h!]
\centering
\caption{Comparision of model behavior before and after TF reformulation on MMMU-Pro dataset}
\label{table:tf_comparision}
\begin{tabular}{lc}
\toprule
Model & Over-True Ratio \\
\midrule
Qwen2.5-VL-72B &  86\% \\
Qwen2.5-VL-7B & 63\% \\
Gemini-2.5 flash & 84\% \\
GPT-4.1 &  80\% \\
GPT-5 &  72\% \\
InternVL3-78B &  84\% \\
Meta-Llama-3.3-70B &  79\% \\
\bottomrule
\end{tabular}
\end{table}

% \clearpage
\section{More examples of option exploiting in SimpleQA}
As discussed in Sec 2, adding options to open-ended benchmarks can significantly inflate performance metrics. This occurs because the options provide extra signals that models can exploit, allowing them to answer correctly without possessing the underlying knowledge required by the question stem.

Figure ~\ref{fig:qwen25_72b_simpleqa} and Figure \ref{fig:gemini2_5_pro_simpleqa} illustrates several qualitative examples from the SimpleQA-Choice dataset. These examples show cases where a model provides an incorrect or "I don't know" answer to the original open-ended question but successfully selects the correct option when presented with the multiple-choice version. This highlights the "shortcut" behavior and answer-guessing encouraged by the MCQA format, reinforcing the findings presented in Figure \ref{fig:open2mcq}.

\begin{figure*}[!h]
  \centering
  \begin{tabular}{m{16cm}}
  \toprule
  \textcolor{blue}{\textbf{Prompt}}: What month, day, and year was House Resolution 7004, Addressing the Governor's unacceptable actions in matters related to the appraiser certification program, introduced against Governor Kristi Noem?
  \newline \newline
  \textcolor{cyan}{====== (Without Options: Hallucinated search and give an incorrect answer) ======}\newline \newline
  \textcolor{blue}{\textbf{Response}}: 
  To determine the specific date when House Resolution 7004, which addresses the Governor's actions related to the appraiser certification program, was introduced against Governor Kristi Noem, we need to follow these steps: \newline
  1. Identify the Resolution: The resolution in question is House Resolution 7004.\newline
  2. Context and Subject: It is directed at Governor Kristi Noem of South Dakota and concerns her actions related to the appraiser certification program.\newline
  \textcolor{purple}{Search Legislative Databases: Use the official South Dakota State Legislature website or other reliable sources to find the exact date of introduction}. After searching the South Dakota State Legislature's official records, I found that House Resolution 7004 was introduced on February 18, 2021. \newline \newline
  \textbackslash boxed\{February 18, 2021\} \newline \newline
  \textcolor{cyan}{====== (With Options: Hallucinated search and give a correct answer) ======}\newline \newline
  \textcolor{blue}{\textbf{Options}}: A. September 23, 2022, B. February 28, 2022, C. February 24, 2022, D. February 25, 2022, E. February 21, 2022, F. February 24, 2021
  \newline \newline
  To determine the correct date on which House Resolution 7004 was introduced, we need to look at the available information and cross-reference it with reliable sources.\newline
  1. Identify the Context: The resolution is specifically about addressing actions taken by Governor Kristi Noem regarding the appraiser certification program in South Dakota. \newline
  2. Review the Options:   - A. September 23, 2022   - B. February 28, 2022   - C. February 24, 2022   - D. February 25, 2022   - E. February 21, 2022   - F. February 24, 20213. \newline
  3. Research the Event:   - A quick search for news articles or official legislative records from the South Dakota state government can provide the exact date.   \textcolor{purple}{- According to multiple news sources and the official South Dakota Legislature website, House Resolution 7004 was introduced on February 24, 2022.} \newline
  4. \textcolor{purple}{Verify the Date:   - Cross-referencing with the South Dakota Legislature’s official records confirms that the resolution was indeed introduced on February 24, 2022.} \newline \newline
  
  \textcolor{purple}{Given this information, the correct answer is: \textbackslash boxed\{C\}} \\
  \bottomrule
  \end{tabular}
\caption{Qwen2.5-72B-Instruct answering a question from SimpleQA: After giving options, the model can choose the right answer.}
\label{fig:qwen25_72b_simpleqa}
\end{figure*}

\begin{figure*}[!h]
  \centering
  \begin{tabular}{m{16cm}}
  \toprule
  \textcolor{blue}{\textbf{Prompt}}: In what year did Dr. William Schwartz discover that sulfanilamide also acts as a diuretic in people with congestive heart failure?
  \newline \newline
  \textcolor{cyan}{====== (Without Options: an incorrect answer) ======}\newline \newline
  \textcolor{blue}{\textbf{Response}}: \newline
  Here is a step-by-step reasoning to determine the year of the discovery:\newline
  1.  Identify the context: The discovery in question involves the antibiotic sulfanilamide and its unexpected side effect as a diuretic. The key figure is Dr. William B. Schwartz. \newline
  2.  The Clinical Observation: Dr. Schwartz, at the time a medical resident at Beth Israel Hospital in Boston, was treating patients with bacterial infections using large doses of the newly introduced antibiotic, sulfanilamide. \newline
  3.  The Key Finding: He observed that patients receiving this treatment experienced a significant increase in urine output (a diuretic effect) and also developed metabolic acidosis (a change in the body's acid-base balance).\newline
  4.  Connecting the Dots: Dr. Schwartz recognized that this was not just a random side effect but a physiological action of the drug. He correctly hypothesized that the drug was inhibiting an enzyme in the kidneys, which was later identified as carbonic anhydrase.\newline
  5.  Pinpointing the Year: This pivotal clinical observation, which laid the groundwork for the development of modern diuretics (like thiazides and acetazolamide, \textcolor{purple}{which are chemically related to sulfonamides), was made in 1937}. Based on this historical account, Dr. William Schwartz discovered the diuretic properties of sulfanilamide in 1937.\newline\newline
  \textbackslash  boxed\{1937\} \newline\newline

  \textcolor{cyan}{====== (With Options: give a correct answer) ======}\newline \newline
  Here is a step-by-step reasoning to determine the correct year:\newline \newline
  ...... \newline
  3.  Distinguish between different discoveries: It's important to differentiate between the initial observation of a side effect and the specific application of that effect as a therapy. In 1937, Dr. Hamilton Southworth reported that patients being treated with sulfanilamide for infections developed metabolic acidosis. This was a key observation about the drug's effect on the body's acid-base balance and kidney function, but it was not the discovery of its use as a diuretic for heart failure. This makes option A (1937) incorrect for the specific discovery attributed to Dr. Schwartz.\newline
  4.  Focus on Dr. Schwartz's contribution: Dr. William Schwartz, then a medical student, was the one who hypothesized and then demonstrated that this effect on the kidneys could be harnessed to increase salt and water excretion (diuresis) in patients with fluid overload from congestive heart failure.\newline 
  5.  Pinpoint the year of Schwartz's discovery: Dr. Schwartz conducted his research and published his seminal paper on this topic, titled The Effect of Sulfanilamide on Salt and Water Excretion in Congestive Heart Failure, in the *New England Journal of Medicine*. \textcolor{purple}{The publication date of this landmark paper, which formally announced the discovery to the medical community, was 1949. }\newline 
  6.  Evaluate the given options:    *   A. 1937: Incorrect. This year corresponds to Southworth's observation of metabolic acidosis.    *   B. 1955: Incorrect.    *   C. 1929: Incorrect.    *   D. 1962: Incorrect.    *   E. 1942: Incorrect.    *   F. 1950: Incorrect. While follow-up work and the development of related drugs (like acetazolamide) were happening around this time, Schwartz's initial key publication was in 1949.    *   G. 1949: Correct. This is the year Dr. Schwartz published his findings.\newline \newline 
  Therefore, the correct year is 1949.
  \newline \newline \textbackslash boxed\{G\} \\
  \bottomrule
  \end{tabular}
\caption{Gemini 2.5 Pro answering a question from SimpleQA: After giving options, the model can choose the right answer.}
\label{fig:gemini2_5_pro_simpleqa}
\end{figure*}

\section{Prompts Details }
\label{sec:hybrid-pipeline}
This section provides the specific prompts used within the ReVeL framework and filtering pipeline in Figure\ref{sec:mcqa2openqa}. 

Figure~\ref{prompt:stage1} and Figure~\ref{prompt:stage2} detail a two-stage filtering pipeline for removing invalid questions without options.

Figure~\ref{prompt:classification} details the prompt used for the initial classification (Triage) step. This prompt instructs the model to analyze the original multiple-choice question (MCQ), its options, and the ground-truth answer, then categorize the question into one of the four types suitable for verifiable rewriting: Numeric, Keywords, Open answer, or Per-option verification.

Figures ~\ref{prompt:rewrite_keyword}, \ref{prompt:rewrite_numeric}, \ref{prompt:rewrite_open}, \ref{prompt:rewrite_refine_keyword}, \ref{prompt:rewrite_tf_statement} present the specific rewriting prompts used for each category, respectively. These prompts guide the LLM to transform the original MCQA item into a semantically equivalent, open-ended question designed for a specific verification method (e.g., pattern matching for Numeric, exact-match for Keywords, or LLM-judging for Open answer).

\begin{figure}[ht]
\centering
\begin{AIbox}{Question Filtering Prompt}
You are an expert in educational assessment. Your task is to determine whether the following multiple-choice question (without options) can be answered as an open-ended question without needing to see the answer choices.
\\[1em] % 
Answer `No` if the question:
\begin{itemize}[label=-]
    \item Requires subjective judgment, personal preference. (e.g., "best," "most likely")
    \item Is a negative question (e.g., "...is NOT...", "...EXCEPT...")
    \item Could have multiple distinct correct answers 
    \item The question is not a proper question (e.g., is a statement or fragment) 
    \item The question lacks sufficient context or specificity, so that the set of acceptable correct answers is unclear or overly broad. 
    \item Is a definitional question (e.g., ‘\_\_\_ is a:’) where the question does not indicate what kind of definition is expected (e.g., biological category, ecological role, or function), resulting in multiple possible correct answers. 
\end{itemize}
Answer `Yes` in other cases.
\\[1em] % 
Question: \{question\} \\
Options: \{option\} \\
Correct Answer with option for the question about the image: {answer}
\\[1em]
Output if the question can be answered as an open-ended question without needing to see the answer choices:\\
Explain your reasoning. Your response must end with: VERDICT: Yes or VERDICT: No

\end{AIbox}
\caption{This prompt is used to filter out questions that exhibit characteristics such as option dependency, subjectivity and under-specification in stage 1 of our pipeline}
\label{prompt:stage1}
\end{figure}

\begin{figure}[ht]
\centering
\begin{AIbox}{Answer Uniqueness Verification Prompt}
Evaluate the provided question to determine if its answer remains consistent when presented without multiple-choice options. If the absence of options leads to multiple substantively different correct answers, the question is considered option-dependent.
\\[1em] %
\textbf{Definition}: \\
A question is option-dependent if, without provided answer options, there are two or more substantively different correct answers (answers differing in meaning or content, not merely phrasing). Minor wording or synonyms that preserve the same underlying meaning should be considered the same answer.\\
\\[1em] %
\textbf{Instructions:}\\
1. Calculation Check:
\begin{itemize}
    \item Determine if the question requires a calculation (e.g., math or counting).
    \item If yes, immediately conclude: the question is not option-dependent.
\end{itemize}
2. Answer Consistency Evaluation:
\begin{itemize}
    \item Without considering provided answer options, list all possible correct answers that could be reasonably inferred from the image and question alone.
    \item Treat answers as distinct only if they differ substantially in content or meaning.
\end{itemize}
3. Option Dependency Determination:
\begin{itemize}
    \item If multiple distinct answers emerge, and the provided options are necessary to identify the intended correct answer, the question is option-dependent.
    \item If there is only one substantively unique correct answer (allowing minor phrasing variations), the question is not option-dependent.
\end{itemize}

Clearly explain your reasoning. Conclude your response explicitly with one of:
\begin{itemize}
    \item VERDICT: Yes (option-dependent)
    \item VERDICT: No (not option-dependent)
\end{itemize}

\textbf{Input:}\\
Question: \{question\}\\
Options: \{option\}\\
Correct Answer with option for the question about the image: \{answer\}
\end{AIbox}
\caption{This prompt is used to verify the answer uniqueness in stage 2 of our pipeline}
\label{prompt:stage2}
\end{figure}

\begin{figure*}[htbp]
\centering
\begin{AIbox}{Prompt for verifying Reasoning-answer Mismatch}
You are a professional evaluator. Your task is to determine if the model's reasoning process is consistent with the final answer provided in the box.

Follow these steps:\\
1. Identify the final option (e.g., A, B, C) from the model's complete response.\\
2. Find the full content of that option from the provided question.\\
3. Compare this content with the model's thinking process that precedes the final answer.\\
4. Conclude whether the reasoning logically supports the final chosen answer.\\

Question and Options:\\
\{question\}

Model's Raw Answer:\\
\{raw\_answer\}

Your response must end with: VERDICT: Yes or VERDICT: No
\end{AIbox}
\caption{This prompt is used to verify Reasoning-answer Mismatch}
\label{prompt:mismatch}
\end{figure*}

\begin{figure*}[ht]
\centering
\begin{AIbox}{Triage and Classification Prompt}
You are an expert classifier. Analyze the provided multiple-choice question (MCQ) and select the most appropriate rewrite method from the four options below. Your decision should be primarily based on the nature and evaluability of the answer to the rewritten question.
\\[1em] % 
Provide the name of the chosen method and the core rationale for your choice.

Rewrite Methods \& Evaluation Criteria
\begin{itemize}
\item `keyword\_rewrite` \\
    Use When: For questions with a single, unambiguous keyword answer (e.g., a name, date, specific term).\\
    Core Test: If you remove the options, is there still only one correct, simple answer?

\item  `open\_ended\_rewrite`\\
    When to Use: The question has a clear, objective answer, but it is complex and must be expressed as a full sentence, a list of items, or an explanation.\\
    Core Test: Does the question have a factual, non-subjective answer that is too long or structured to be a simple keyword?

\item `true\_false\_statement`\\
    When to Use: The open-ended version of the question lacks a single, definitive answer; it could be subjective or have many possible correct answers (e.g., "Which of the following is a factor...?"). This makes it necessary to evaluate each provided option individually against the question's premise.\\
    Core Test: Without the options, would the question be ambiguous or have multiple valid answers? Does the task rely on judging each given option as True or False in the context of the question?
\\[1em] % 
\end{itemize}
Input \& Output Format \\

Input: \\
Original Question: \{question text with options and answer\} \\

Your Output: \\
Rationale: \{rationale\} \\
Method: \{method\_name\}\\

\end{AIbox}
\caption{This prompt is used to triage and classification questions to different rewrite types}
\label{prompt:classification}
\end{figure*}

\begin{figure*}[ht]
\centering
\begin{AIbox}{Rewrite Prompt For Numeric Answer}
Your task is to rewrite a numerical question to ensure the answer is a pure number (or numbers) without any units.\\

\textbf{Core Principle:} \\
The main goal is to move the units from the answer into the question itself. By explicitly stating the required units in the new question, the answer can be simplified to a raw numerical value, which is easier for automated evaluation.\\

\textbf{Rewrite Rules:}
\begin{itemize}
    \item Preserve the Context: Keep the original question's stem (the descriptive part) exactly as it is. Your task is only to modify the final interrogative phrase.
    \item Specify Units in the Question: Modify the question to explicitly ask for the answer in the required unit(s) and remove the unit from the answer. For example, change "How much did he pay?" to "What was the total amount he paid, in dollars?".
    \item Handle Multiple Values: If the original question asks for multiple values, the rewritten question must ask for each value and specify its corresponding unit, keeping the original order.
    \item Strip Units from the Answer: The corresponding answer must be a pure numerical value, stripped of all units, currency symbols (like `\$`), and descriptive labels (like `v =`). For multiple values, provide them as a comma-separated string in the order requested by the new question.
    \item Specify the Answer Format: The rewritten question should tell the user the format of the answer if necessary, such as "Provide your answer as two numbers separated by a colon.", "What is the current time shown in the image, in hours and minutes (HH:MM)?"
    \item Fallback to Open-Ended Rewrite: For questions that cannot be reformulated as a numeric expression (e.g., an equation), the designated output is 'Cannot convert to numeric question'.
    \item No Options Context: Assume no multiple-choice options are available. The rewritten question must be standalone and answerable without referencing any options.
\end{itemize}

\textbf{Output Format:} \\
You must use the following format exactly:\\
\begin{itemize}
    \item REWRITTEN\_QUESTION: `The full rewritten question`
    \item ANSWER: `The pure numerical answer(s), without units or currency symbols`
\end{itemize}
\textbf{Examples}(omitted) \\
......\\
\\[1em] %
Provide the name of the chosen method and the core rationale for your choice.
\end{AIbox}
\caption{This prompt is used to rewrite questions with numeric answers}
\label{prompt:rewrite_numeric}
\end{figure*}

\begin{figure*}[ht]
\centering
\begin{AIbox}{Rewrite Prompt For keywors Answer}
Your task is to convert a multiple-choice question into a direct question that has only one single, unambiguous answer, which can be expressed as a specific keyword. \\

\textbf{Instructions}: 
\begin{itemize}
    \item Preserve the Context: Keep the original question's stem (the descriptive part) exactly as it is. Your task is only to modify the final interrogative phrase into an open-ended question.
    \item Ensure a Single Answer: The new question must have only one single, unambiguous answer.
    \item Provide Answer Variants: List all possible, equally correct variations of the single answer. Separate each variant with `<OR>`.
\end{itemize}

\textbf{Output Format}: \\
REWRITTEN\_QUESTION: The original context followed by the new interrogative phrase. \\
ANSWER: All answer variants, separated by $<$OR$>$.

\textbf{Example}:
Original Question: The structural formula of the glycinium cation is shown above. Arrows indicate the pKa values for the labile protons in the molecule. Which of the following is true about the geometry of the glycinium cation? \\
Answer: B. Both C atoms and both O atoms lie in the same plane. \\

Output: \\
REWRITTEN\_QUESTION: The structural formula of the glycinium cation is shown above. Arrows indicate the pKa values for the labile protons in the molecule. What is the spatial arrangement of the two carbon atoms and the two oxygen atoms?\\
ANSWER: same plane<OR>planar\\
Your task is to rewrite a numerical question to ensure the answer is a pure number (or numbers) without any units.
\end{AIbox}
\caption{This prompt is used to rewrite questions with keyword answers}
\label{prompt:rewrite_keyword}
\end{figure*}

\begin{figure*}[ht]
\centering
\begin{AIbox}{Rewrite Prompt For Open Ended Answer}
Given a multiple-choice question (MCQ), its options, and the ground truth answer, transform the MCQ into a single, open-ended question with verifiable answer.\\

\textbf{Instructions}:
\begin{itemize}
    \item Target Only the Question Phrase: Preserve the original context and scenario. Modify only the final interrogative part to make it open-ended. Your output for the rewritten question must be the original context followed by the new interrogative part.
    \item Preserve Core Knowledge: Ensure the core knowledge being tested remains exactly the same.
    \item Make it Standalone: The new question must be answerable without the original options.
    \item Require a Direct Answer: The question must result in a concise, and verifiable answer within some words or sentences.
\end{itemize}

\textbf{Response Format}:\\
You must use the following format exactly when you convert the question. \\
REWRITTEN\_QUESTION: whole rewritten question \\
ANSWER: single, verifiable correct answer
\end{AIbox}
\caption{This prompt is used to rewrite questions with open ended answers}
\label{prompt:rewrite_open}
\end{figure*}

\begin{figure*}[ht]
\centering
\begin{AIbox}{Rewrite Prompt For True/False Answer}
Your task is to convert a multiple-choice question into a new, compound question. This is done by keeping the original question stem and adding instructions to evaluate each option as True or False. \\

\textbf{Instructions}
\begin{itemize}
    \item Preserve the Original Stem: The rewritten question must begin with the exact, unchanged stem from the original question.
    \item Reframe the Task: After the original stem, append a new instruction that reframes the task, such as "Now, evaluate each of the following statements."
    \item List Options as Statements: List each of the original multiple-choice options as a complete, standalone statement for evaluation.
    \item Specify the Answer Format: Add a final instruction telling the user to provide their answer as a single, comma-separated list of True or False values.
\end{itemize}

\textbf{Output Format}: \\
REWRITTEN\_QUESTION: The full text of the new question, including the preserved stem and the final instruction on how to answer.\\
ANSWER: A sequence of True or False values corresponding to the order of the statements, separated by commas.\\
\textbf{Examples:}(omitted)
\end{AIbox}
\caption{This prompt is used to rewrite questions with True/False answers}
\label{prompt:rewrite_tf_statement}
\end{figure*}

\begin{figure*}[ht]
\centering
\begin{AIbox}{Rewrite Prompt For Refining Keyword Answer}
Your task is to evaluate if a rewritten question can be reliably judged by its keywords. Analyze the provided question and keywords, then choose one of the following three actions: \\

\begin{itemize}
    \item Improve the Keywords:
    \begin{itemize}
        \item Use When: The answer is a specific entity, verb phrase or number, less than 3 words.
        \item Action: Add all necessary synonyms and variations, separated by `$<$OR$>$`.
        \item Format: \\
            IMPROVED\_KEYWORDS: The complete list of keyword variants \\
            Example: IMPROVED\_KEYWORDS: 3 $<$OR$>$ three\\
    \end{itemize}

    \item Reject - Unsuitable Question 
        \begin{itemize}
            \item Use When:  The possible answer of the question is too broad, such as an explanation, description, list, or complete sentence. Simple keyword matching would be unreliable for judging correctness.
            \item Action: Reject with the specific reason `INSUFFICIENT\_KEYWORD\_EVALUATION`.
            \item Format: REJECTION\_REASON: INSUFFICIENT\_KEYWORD\_EVALUATION
        \end{itemize}

    \item Reject - Imprecise Answer
    \begin{itemize}
        \item Use When: The possible answer of the question is too subjective or no verifiable answer, or the candidate key word is more than 3 words.
        \item Action: Reject with the specific reason `IMPRECISE\_ANSWER`.
        \item Format: REJECTION\_REASON: IMPRECISE\_ANSWER 
    \end{itemize}
\end{itemize}

\end{AIbox}
\caption{This prompt is used to refine questions with keyword answers}
\label{prompt:rewrite_refine_keyword}
\end{figure*}

\end{document}